\RequirePackage{amsmath}
\documentclass[runningheads]{llncs}
\usepackage[T1]{fontenc}
\usepackage{geometry}
\usepackage{amssymb}
\usepackage{multirow}
\usepackage{graphicx,verbatim}
\usepackage{subfigure}
\usepackage{float}
\usepackage{hyperref}
\hypersetup{
    colorlinks=true,
    linkcolor=blue,
    citecolor=magenta,
    urlcolor=blue
}
\usepackage{color}

\begin{document}
\title{ODySSeI: \\An Open-Source End-to-End Framework for Automated Detection, Segmentation, and Severity Estimation of Lesions in \\Invasive Coronary Angiography Images}
\author{Anand Choudhary\inst{1} \and Xiaowu Sun\inst{1,2,6} \and Thabo Mahendiran\inst{4} \and Ortal Senouf\inst{1,2} \and Denise Auberson\inst{4} \and Bernard De Bruyne\inst{4,5} \and Stephane Fournier\inst{4} \and Olivier Muller\inst{4} \and Emmanuel Abbé\inst{2,3} \and Pascal Frossard\inst{1,3} \and
Dorina Thanou\inst{1,3}}
\institute{LTS4 Laboratory, EPFL, Lausanne, Switzerland \\
\and
Chair of Mathematical Data Science, EPFL, Lausanne, Switzerland \\
\and
EPFL AI Center, Lausanne, Switzerland \\
\and
Cardiology Department, CHUV, Lausanne, Switzerland\\
\and
Cardiovascular Center, OLV, Aalst, Belgium\\
\and 
School of AIAC, Xi'an Jiaotong-Liverpool University, Suzhou, China\\
\email{anand.choudhary@alumni.epfl.ch, dorina.thanou@epfl.ch}
}
\maketitle              
\begin{abstract}
Invasive Coronary Angiography (ICA) is the clinical gold standard for the assessment of coronary artery disease. However, its interpretation remains subjective and prone to intra- and inter-operator variability. In this work, we introduce \textbf{ODySSeI}: an \textbf{O}pen-source end-to-end framework for automated \textbf{D}etection, \textbf{S}egmentation, and \textbf{S}everity estimation of lesions in \textbf{I}CA images. ODySSeI integrates deep learning-based lesion detection and lesion segmentation models trained using a novel Pyramidal Augmentation Scheme (PAS) to enhance robustness and real-time performance across diverse patient cohorts (2149 patients from Europe, North America, and Asia). Furthermore, we propose a quantitative coronary angiography-free Lesion Severity Estimation (LSE) technique that directly computes the Minimum Lumen Diameter (MLD) and diameter stenosis from the predicted lesion geometry. Extensive evaluation on both in-distribution and out-of-distribution clinical datasets demonstrates ODySSeI's strong generalizability. Our PAS yields large performance gains in highly complex tasks as compared to relatively simpler ones, notably, a 2.5-fold increase in lesion detection performance versus a 1-3\% increase in lesion segmentation performance over their respective baselines. Our LSE technique achieves high accuracy, with predicted MLD values differing by only 2–3 pixels from the corresponding ground truths. On average, ODySSeI processes a raw ICA image within only a few seconds on a CPU and in a fraction of a second on a GPU and is available as a plug-and-play web interface at \href{https://swisscardia.epfl.ch}{swisscardia.epfl.ch}. Overall, this work establishes ODySSeI as a comprehensive and open-source framework which supports automated, reproducible, and scalable ICA analysis for real-time clinical decision-making.

\end{abstract}

\section{Introduction}
Cardiovascular Diseases (CVDs) accounted for more than 19 million deaths worldwide in 2021~\cite{AHA_update}. Coronary Artery Disease (CAD) is the leading cause of CVD-related deaths, with its mortality rate increasing by approximately 72\% from 1990 to 2021~\cite{AHA_update}. Therefore, early diagnosis of CAD is critical for initiating timely therapeutic interventions, preventing adverse cardiac events, and improving patient prognosis.

CAD refers to heart problems caused by reduced blood supply (ischemia) secondary to the narrowing of the coronary artery lumen by atherosclerotic lesions~\cite{IHD}. Invasive Coronary Angiography (ICA) is the clinical gold standard for the anatomical assessment of coronary artery stenosis~\cite{CAR_guidelines}.

However, the interpretation of ICA is inherently subjective, and highly dependent on operator experience. Intra- and inter-operator variability have been reported to reach up to 45\%~\cite{ICA_variability}, potentially leading to inconsistent clinical decision-making. Although Quantitative Coronary Angiography (QCA) can provide a more objective assessment of coronary artery stenosis, it relies heavily on manual lesion identification and is time-consuming, which limits its routine use in clinical practice~\cite{QCA,angiopy}. There is thus, an important, unmet clinical need for automated end-to-end frameworks that can support the objective and accurate quantitative assessment of lesions, directly from ICA images in real-time.

Deep learning has demonstrated promising levels of performance in lesion detection and lesion segmentation in ICA images~\cite{lesion_det2,lesion_det3,StenDet,ARCADE,arcade_2}. However, the reliance on regional datasets and dataset-specific post-processing limits the generalizability of previous deep learning-based methods to real-world clinical settings. Moreover, these methods
assume the presence of lesions in all the input images, introducing an undesirable bias that makes them non-conforming to real-world clinical distributions. More recent end-to-end pipelines such as CathAI~\cite{CathAI} and DeepCoro~\cite{DeepCoro} integrate 5~\cite{CathAI} and 12 deep learning models~\cite{DeepCoro} (respectively), leading to high computational latencies and limited real-time applicability. Moreover, these pipelines struggle to detect severely narrowed, often critical lesions which further constrains their translational utility~\cite{CathAI,DeepCoro}.

\begin{figure}[!htb]
    \begin{center}
        \subfigure[Inference-time operation of ODySSeI]{
            \label{fig_in_act}
            \includegraphics[scale = 0.45]{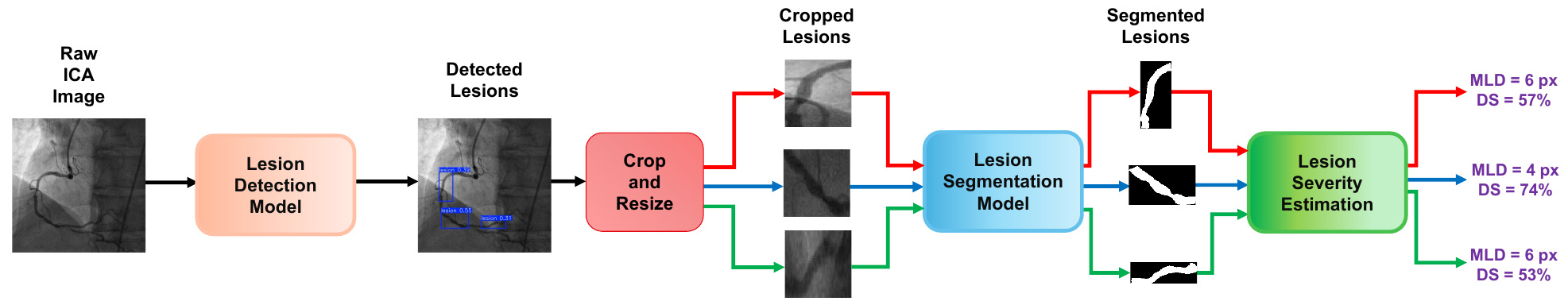}
        }
        \subfigure[Fine-tuning ODySSeI's lesion detection model]{
            \label{fig_ft_ld}
            \includegraphics[scale = 0.28]{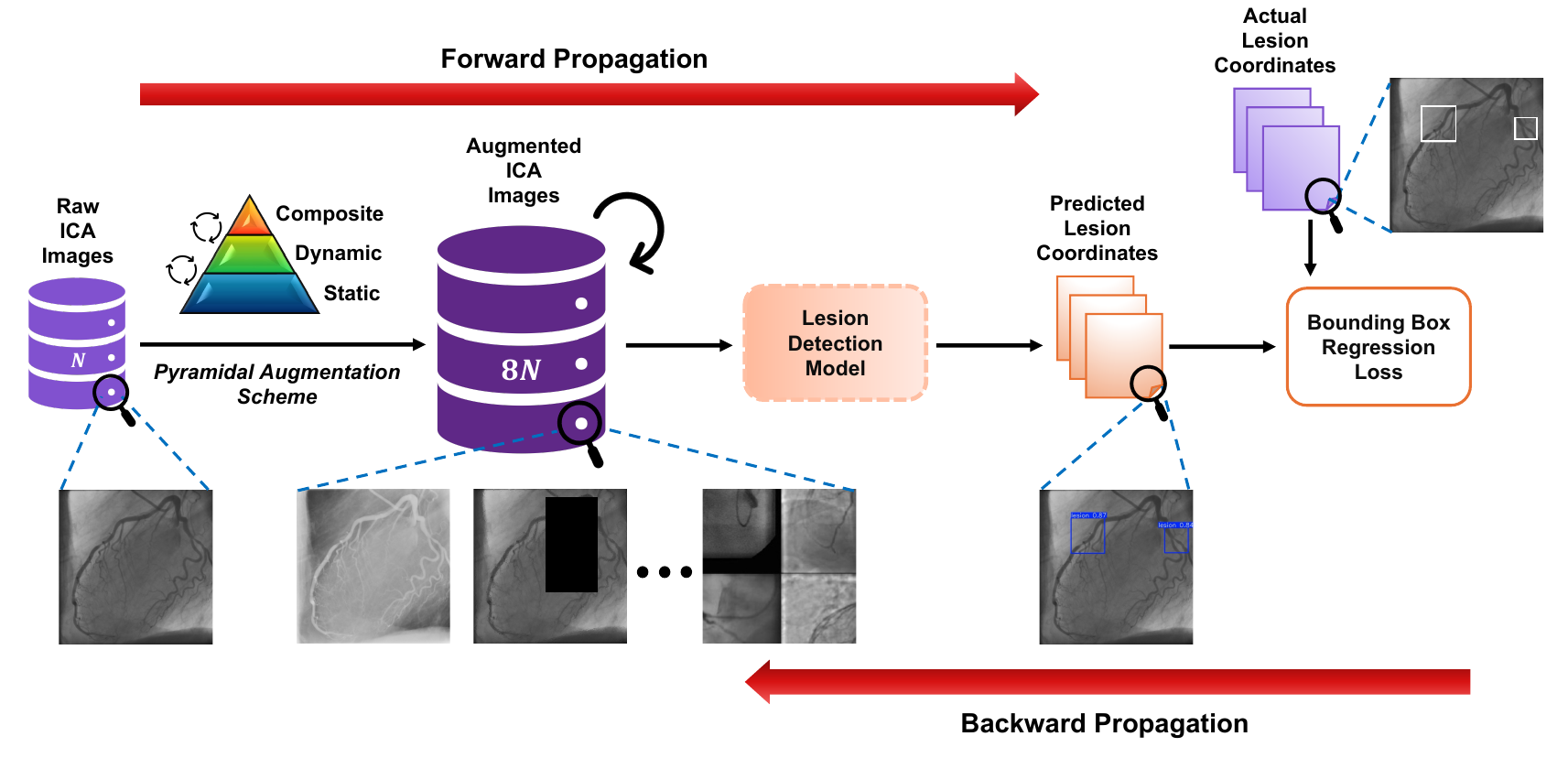}
        }
        \subfigure[Fine-tuning ODySSeI's lesion segmentation model]{
            \label{fig_ft_ls}
            \includegraphics[scale = 0.28]{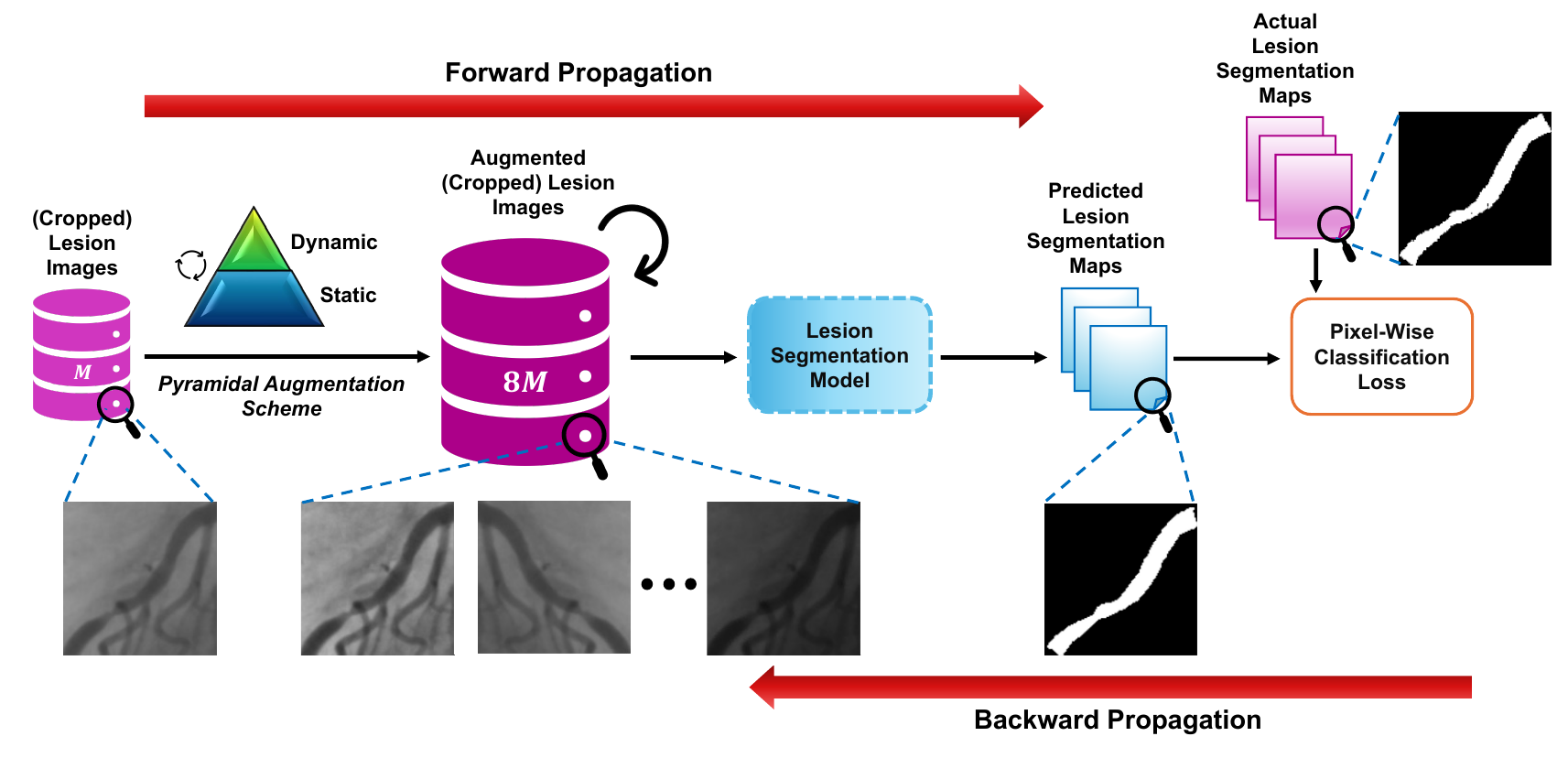}
        }
        \subfigure[Pyramidal Augmentation Scheme (PAS)]{
            \label{pas}
            \includegraphics[scale = 0.30]{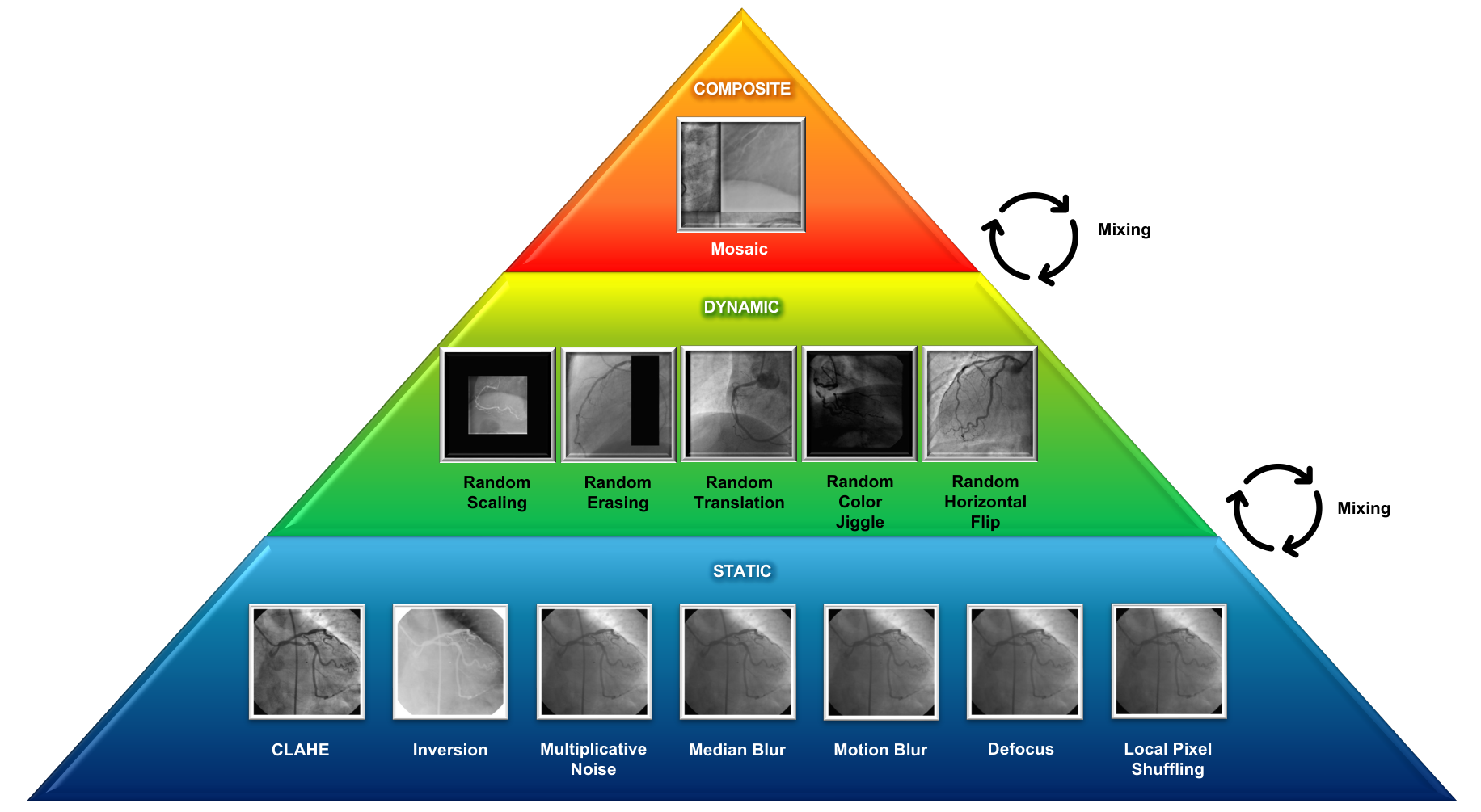}
        }
        \subfigure[Lesion severity estimation technique]{
            \label{lse}
            \includegraphics[scale = 0.32]{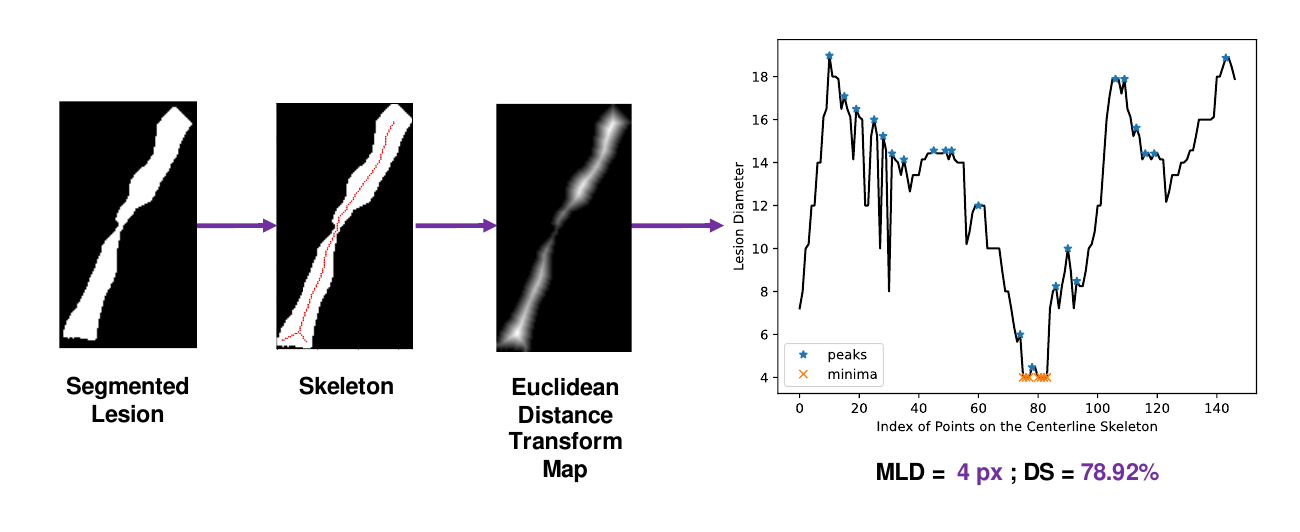}
        }
    \end{center}
    \caption{\textbf{Overview of ODySSeI}. (a) During inference, a raw ICA image undergoes detection, segmentation (after cropping and resizing the lesion instances detected in the image), and severity estimation of any lesions present in it. (b) Raw training data is constantly augmented using all the three tiers of our Pyramidal Augmentation Scheme (PAS) to iteratively fine-tune our lesion detection model for the identification of lesions and prediction of their bounding box coordinates. The fine-tuned lesion detection model is then integrated into ODySSeI for inference. (c) Cropped training data (only lesion instances) is constantly augmented using two tiers of our PAS to iteratively fine-tune our lesion segmentation model for the prediction of lesion geometry. The fine-tuned lesion segmentation model is then integrated into ODySSeI for inference. (d) The three tiers of our PAS: static, dynamic, and composite augmentations. (e) Our lesion severity estimation technique involves computing the arterial radius of every point on the centerline skeleton of the segmented lesion. Twice of the minimum between two radii peaks yields the Minimum Lumen Diameter (MLD). Complement of the ratio of the MLD to the highest peak's diameter yields the Diameter Stenosis (DS).}
    \label{end2end_full}
\end{figure}

To address these challenges, we present \textbf{ODySSeI}: an \textbf{O}pen-source end-to-end framework for automated \textbf{D}etection, \textbf{S}egmentation, and \textbf{S}everity estimation of lesions in \textbf{I}CA images. As illustrated in~\autoref{fig_in_act}, ODySSeI first applies a lesion detection model to localize candidate lesions (if present). Detected lesions are then cropped and resized to serve as the input of a lesion segmentation model which delineates lesion geometry. Based on the predicted segmentation maps, a fully automated, novel, QCA-free algorithm then computes clinically relevant lesion severity metrics including the Minimum Lumen Diameter (MLD) and Diameter Stenosis (DS).

ODySSeI is trained using a novel Pyramidal Augmentation Scheme (PAS) designed to improve robustness and generalizability. Our PAS integrates three types of augmentations: seven domain-specific \textit{static} augmentations, which increase dataset size and enforce the learning of arterial geometry and topology while negating reliance on shortcuts such as noise and contrast; five probabilistic \textit{dynamic} augmentations, which improve robustness to plausible imaging variations; and a scene-based \textit{composite} augmentation, which simulates occlusion and promotes compositional learning. Our PAS is used for training both our lesion detection (see~\autoref{fig_ft_ld}) and lesion segmentation (see~\autoref{fig_ft_ls}) models to make them clinically robust and performant in real-time.

Performance evaluations across clinically diverse cohorts from Europe, North America, and Asia demonstrate the accuracy and generalizability of ODySSeI. Furthermore, ODySSeI processes a raw ICA image within only a few seconds on a CPU and in a fraction of a second on a GPU, indicating its high computational efficiency for real-time clinical applications. Combined with its plug-and-play nature, all these benefits make ODySSeI a valuable and user-friendly tool for clinical decision-making. ODySSeI is open-source and live at \href{https://swisscardia.epfl.ch}{swisscardia.epfl.ch} and its associated code is available at \href{https://github.com/LTS4/ODySSeI}{github.com/LTS4/ODySSeI}.

\section{Results}

\subsection{Datasets}

We use three different datasets for training and/or testing ODySSeI. We use the FAME2 dataset~\cite{FAME2_orig} for training and evaluating ODySSeI's lesion detection and lesion segmentation models and the publicly available ARCADE dataset~\cite{ARCADE} for training ODySSeI's lesion detection model. Furthermore, the Future Culprit (FC) dataset~\cite{Future_Culprit} forms an Out-Of-Distribution (OOD) evaluation dataset since it is not used for training ODySSeI's models and comes from a different patient cohort than the FAME2 and ARCADE datasets. Further details pertaining to these datasets are provided in~\autoref{datasets}.

In the following subsections, we provide the results of all our experiments on lesion detection (\autoref{ldet_res_full}), lesion segmentation (\autoref{lseg_res}), and lesion severity estimation (\autoref{lse_res}). Unless otherwise stated, all our results correspond to FAME2's training, validation, and test sets. Abbreviations used for all the evaluation metrics in the subsequent tables and figures are as mentioned in~\autoref{sec:eval_metrics}.

\subsection{Lesion Detection Performance}

In this subsection, we first report the results based on conventional detection metrics which evaluate the overlap between the predicted and ground truth bounding boxes for lesions (\autoref{ldet_res}). We then present the results using our proposed metrics which are defined on the basis of whether the predicted lesion bounding box contains the ground truth MLD (\autoref{mld_based_ldet_res}).

\label{ldet_res_full}
\subsubsection{Bounding Box Overlap-Based Evaluation}
\label{ldet_res}
\paragraph{\textbf{Validation Set Performance}}
\label{init_ldet_res}
\begin{table}[H]
\centering
\caption{\textbf{PAS-based data augmentation substantially boosts ODySSeI's lesion detection performance}. The static augmentation tier consists of seven domain-specific augmentations, the dynamic augmentation tier consists of five probabilistic configurational augmentations, and the composite augmentation tier consists of a scene-based augmentation. Note that the `none' augmentation suite refers to the baseline model which is trained without any data augmentations.}
\resizebox{\textwidth}{!}{%
    \begin{tabular}{|c|c|c|c|c|c|c|c|c|}
        \hline
        \textbf{}&\multicolumn{8}{|c|}{\textbf{Validation Set Metrics}}\\
        \cline{2-9}
        \textbf{Augmentation Suite}&\multicolumn{4}{|c|}{\textbf{Image-Level$^a$}}&\multicolumn{4}{|c|}{\textbf{Lesion-Level$^b$}}\\
        \cline{2-9}
        \textbf{}&\textbf{Prec$^c$}&\textbf{Rec}&\textbf{mAP@0.50$^c$}&\textbf{mAP@0.50-0.95$^c$}&\textbf{Prec}&\textbf{Rec}&\textbf{mAP@0.50}&\textbf{mAP@0.50-0.95}\\
        \hline
        \textbf{None}&\text{$0.673 \pm 0.326$}&\text{$0.401 \pm 0.474$}&\text{$0.324 \pm 0.721$}&\text{$0.287 \pm 0.363$}&\text{$0.342$}&0.221&\text{$0.175$}&\text{$0.073$}\\
        \hline
        \textbf{Static}&$\mathbf{1.000 \pm 0.000}$&\text{$0.061 \pm 0.240$}&$\mathbf{0.995 \pm 0.000}$&$\mathbf{0.597 \pm 0.081}$&0.445&\text{$0.052$}&0.241&0.145\\	
        \hline						
        \textbf{Static + Dynamic}&\text{$0.545 \pm 0.310$}&$\text{$0.857 \pm 0.299$}$&\text{$0.693 \pm 0.346$}&\text{$0.349 \pm 0.226$}&\text{$0.383$}&$\mathbf{0.454}$&\text{$0.374$}&\text{$0.150$}\\	
        \hline
        \textbf{Dynamic + Composite}&\text{$0.410 \pm 0.355$}&$\mathbf{0.899 \pm 0.256}$&\text{$0.618 \pm 0.376$}&\text{$0.325 \pm 0.258$}&\text{$0.442$}&\text{$0.338$}&\text{$0.337$}&\text{$0.146$}\\	
        \hline
        \textbf{Static + Dynamic + Composite}&\text{$0.713 \pm 0.299$}&\text{$0.661 \pm 0.447$}&\text{$0.789 \pm 0.277$}&\text{$0.399 \pm 0.270$}&$\mathbf{0.455}$&\text{$0.390$}&$\mathbf{0.433}$&$\mathbf{0.192}$\\
        \hline
        \multicolumn{6}{l}{\footnotesize{$^a$Mean $\pm$ standard deviation are reported over per-image metrics;}}\\
        \multicolumn{6}{l}{\footnotesize{$^b$Total values are reported (all instances of lesions across all images are considered together);}}\\
        \multicolumn{6}{l}{\footnotesize{$^c$Mean $\pm$ standard deviation are only reported over those images for which the model detects at least one lesion}}\\
        \multicolumn{6}{l}{\footnotesize{ since precision is undefined otherwise. Further, standard deviation values are expected to be high since the model can }}\\ 
        \multicolumn{6}{l}{\footnotesize{ predict FNs. Accordingly, lesion-level metrics, which account for FPs and FNs in total, provide more accurate results.}}\\
    \end{tabular}%
    }
    \label{pyram_aug_det_full}
\end{table}
\begin{table}[H]
\centering
\caption{\textbf{Addition of an external patient cohort (ARCADE dataset) together with PAS-based data augmentation increases ODySSeI's lesion detection performance}. The AUG\textsuperscript{S}-FAME2 dataset refers to the statically augmented FAME2 dataset. Dynamic and composite augmentations are always applied during training, on all datasets.}
\resizebox{\textwidth}{!}{%
    \begin{tabular}{|c|c|c|c|c|c|c|c|c|}
        \hline
        \textbf{}&\multicolumn{8}{|c|}{\textbf{Validation Set Metrics}}\\
        \cline{2-9}
        \textbf{Training Datasets}&\multicolumn{4}{|c|}{\textbf{Image-Level$^a$}}&\multicolumn{4}{|c|}{\textbf{Lesion-Level$^b$}}\\
        \cline{2-9}
        \textbf{}&\textbf{Prec$^c$}&\textbf{Rec}&\textbf{mAP@0.50$^c$}&\textbf{mAP@0.50-0.95$^c$}&\textbf{Prec}&\textbf{Rec}&\textbf{mAP@0.50}&\textbf{mAP@0.50-0.95}\\
        \hline	
        \textbf{AUG\textsuperscript{S}-FAME2}&$\mathbf{0.713 \pm 0.299}$&\text{$0.661 \pm 0.447$}&$\mathbf{0.789 \pm 0.277}$&\text{$0.399 \pm 0.270$}&\text{$0.455$}&\text{$0.390$}&\text{$0.433$}&\text{$0.192$}\\
        \hline
        \textbf{FAME2 x ARCADE}&\text{$0.502 \pm 0.329$}&$\mathbf{0.937 \pm 0.188}$&\text{$0.363 \pm 0.655$}&\text{$0.223 \pm 0.331$}&\text{$0.483$}&$\mathbf{0.480}$&\text{$0.438$}&\text{$0.161$}\\	
        \hline
        \textbf{AUG\textsuperscript{S}-FAME2 x ARCADE}&\text{$0.623 \pm 0.353$}&\text{$0.843 \pm 0.327$}&\text{$0.712 \pm 0.351$}&$\mathbf{0.393 \pm 0.265}$&$\mathbf{0.601}$&\text{$0.431$}&$\mathbf{0.451}$&$\mathbf{0.211}$\\
        \hline
        \multicolumn{6}{l}{\footnotesize{$^a$Mean $\pm$ standard deviation are reported over per-image metrics;}}\\
        \multicolumn{6}{l}{\footnotesize{$^b$Total values are reported (all instances of lesions across all images are considered together);}}\\
        \multicolumn{6}{l}{\footnotesize{$^c$Mean $\pm$ standard deviation are only reported over those images for which the model detects at least one lesion}}\\
        \multicolumn{6}{l}{\footnotesize{ since precision is undefined otherwise. Further, standard deviation values are expected to be high since the model can }}\\ 
        \multicolumn{6}{l}{\footnotesize{ predict FNs. Accordingly, lesion-level metrics, which account for FPs and FNs in total, provide more accurate results.}}\\
    \end{tabular}%
    }
    \label{fameXarcade}
\end{table}
\begin{figure*}[!htb]
    \begin{center}
        \subfigure[]{
            \label{fig_ldld_a}
            \includegraphics[scale = 0.35]{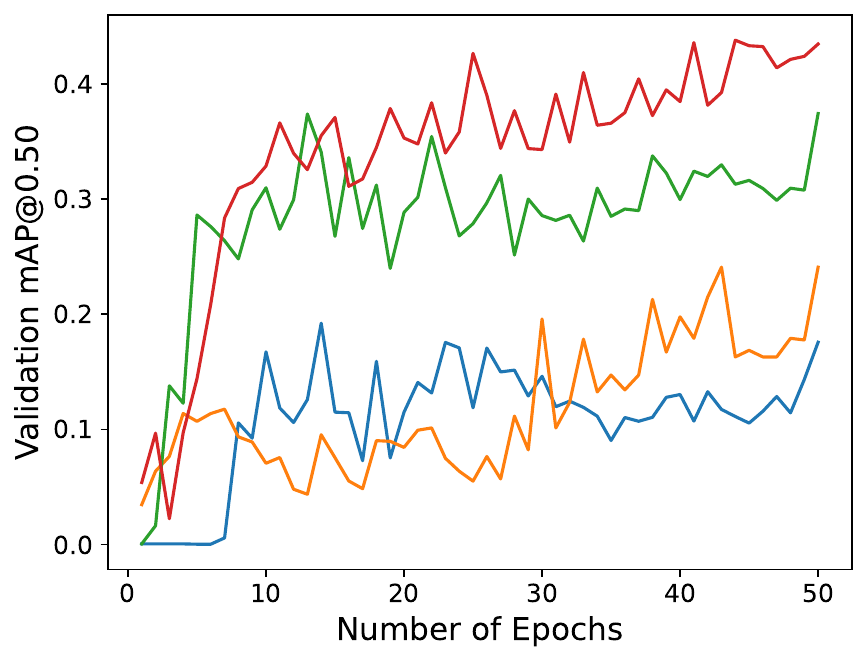}
        }
        \subfigure[]{
            \label{fig_ldld_b}
            \includegraphics[scale = 0.35]{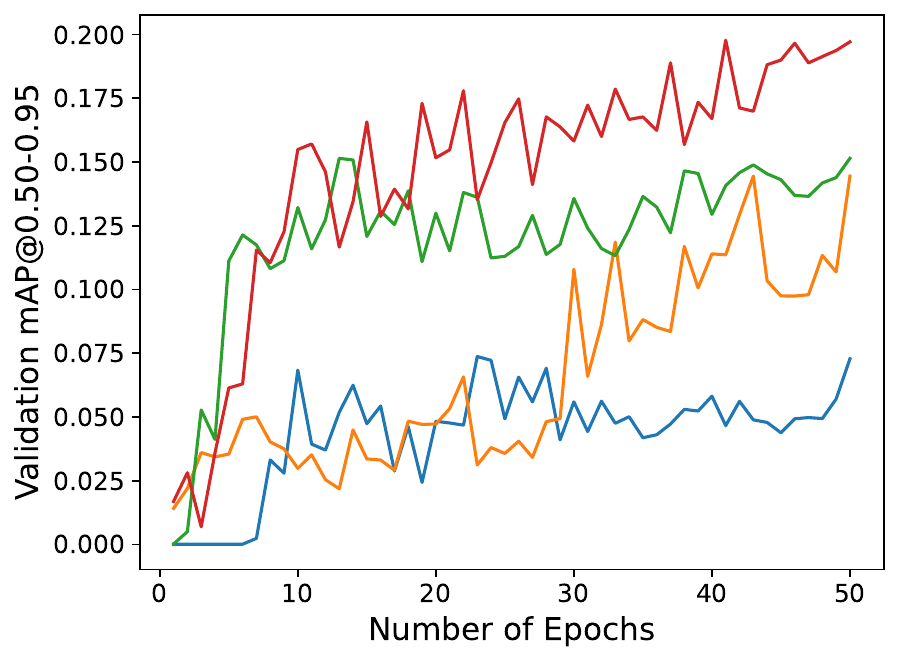}
        }
        \subfigure[]{
            \label{fig_ldld_c}
            \includegraphics[scale = 0.35]{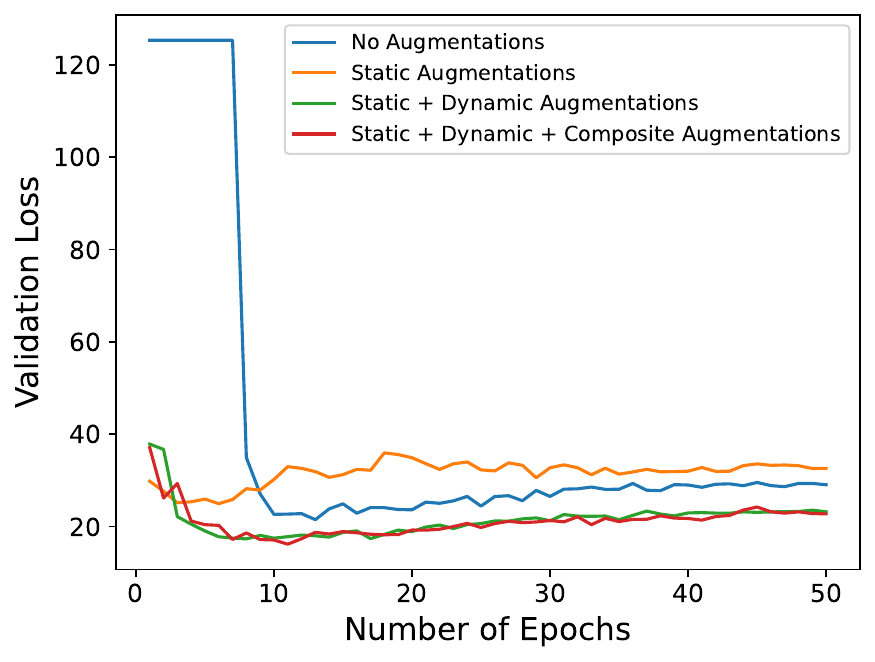}
        }
    \end{center}
    \caption{\textbf{PAS-based data augmentation changes the learning dynamics of ODySSeI's lesion detection model} and encourages structured overfitting via static augmentations, imparts robustness and stabilization via dynamic augmentations, and improves precision in lesion localization via composite augmentations. Plots show (a) lesion-level mAP@0.50 on the FAME2 validation set, (b) lesion-level mAP@0.50-0.95 on the FAME2 validation set, and (c) loss on the FAME2 validation set.}
    \label{fig_pas_LD_learning_dynamics}
\end{figure*}
\autoref{pyram_aug_det_full} shows the effect of our PAS on the performance of our lesion detection model, which is based on the YOLO11 architecture~\cite{yolo11}. Clearly, the baseline model, i.e., the model trained on the non-augmented FAME2 training set, fails to accurately localize lesions since it only manages to achieve an mAP@0.50 of 0.175 at the lesion-level. Static augmentations increase the precision but decrease the recall. Dynamic augmentations mitigate this issue and improve the localization generalizability by counteracting overfitting (as evidenced by the substantially increased value of recall at both the lesion-level and the image-level). Composite augmentations further increase the mAP@0.50 by 6\% at the lesion-level (total) and by 17.1\% at the image-level (on average) by promoting the learning of complex spatial compositions (compare the 3\textsuperscript{rd} and the 5\textsuperscript{th} row of~\autoref{pyram_aug_det_full}). Although static augmentations might seem redundant, the last two rows of~\autoref{pyram_aug_det_full} prove that overfitting to diverse real-life ICA scenarios, i.e., \textit{structured} overfitting, also aids in generalizability. These performance improvements are also in line with the changes in our model's learning dynamics, as affected by our PAS (see~\autoref{fig_pas_LD_learning_dynamics}): static augmentations encourage \emph{structured} overfitting, dynamic augmentations impart robustness and performance stability, and composite augmentations increase the precision in lesion localization.

Overall, our PAS yields a 2.5-fold improvement in mAP@0.50 over the baseline model at both the lesion-level (total) and image-level (on average). Furthermore, addition of an external patient cohort, i.e., ARCADE's training set, to the statically augmented FAME2 (AUG\textsuperscript{S}-FAME2) training set (not just FAME2) along with dynamic and composite augmentations results in the final best lesion detection model (see~\autoref{fameXarcade}). Hence, augmentation of training data by firstly, using our PAS (synthetic augmentation of existing data) and secondly, adding an external patient cohort (real augmentation with new data) leads to large performance gains in lesion detection.

\begin{figure*}[!htb]
    \begin{center}
        \subfigure[]{
            \label{figfva}
            \includegraphics[scale = 0.12]{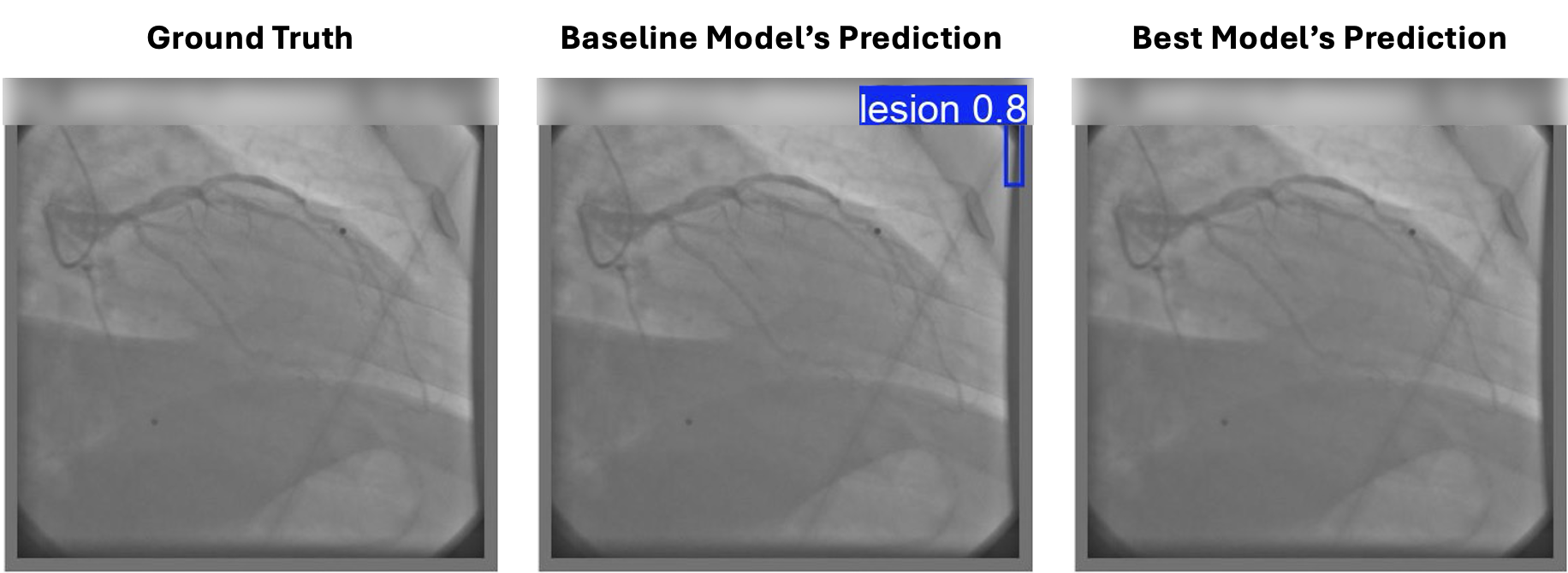}
        }
        \subfigure[]{
            \label{figfvb}
            \includegraphics[scale = 0.12]{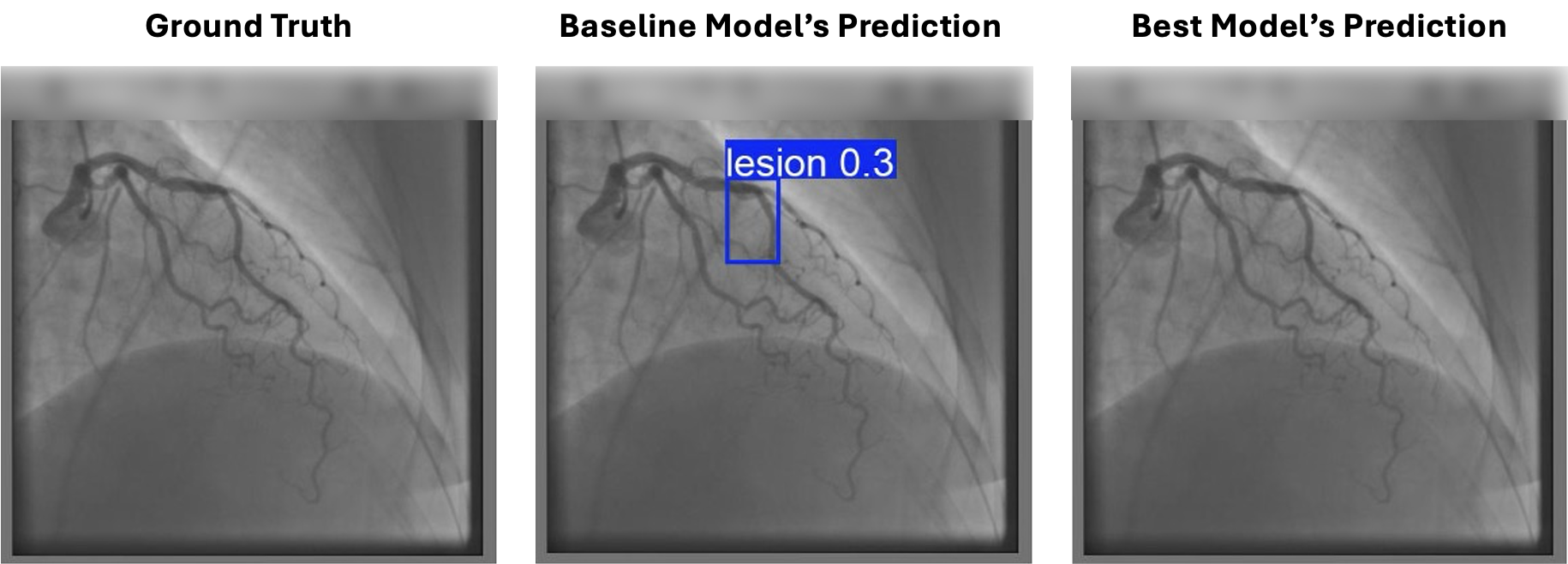}
        }
        \subfigure[]{
            \label{figfvc}
            \includegraphics[scale = 0.12]{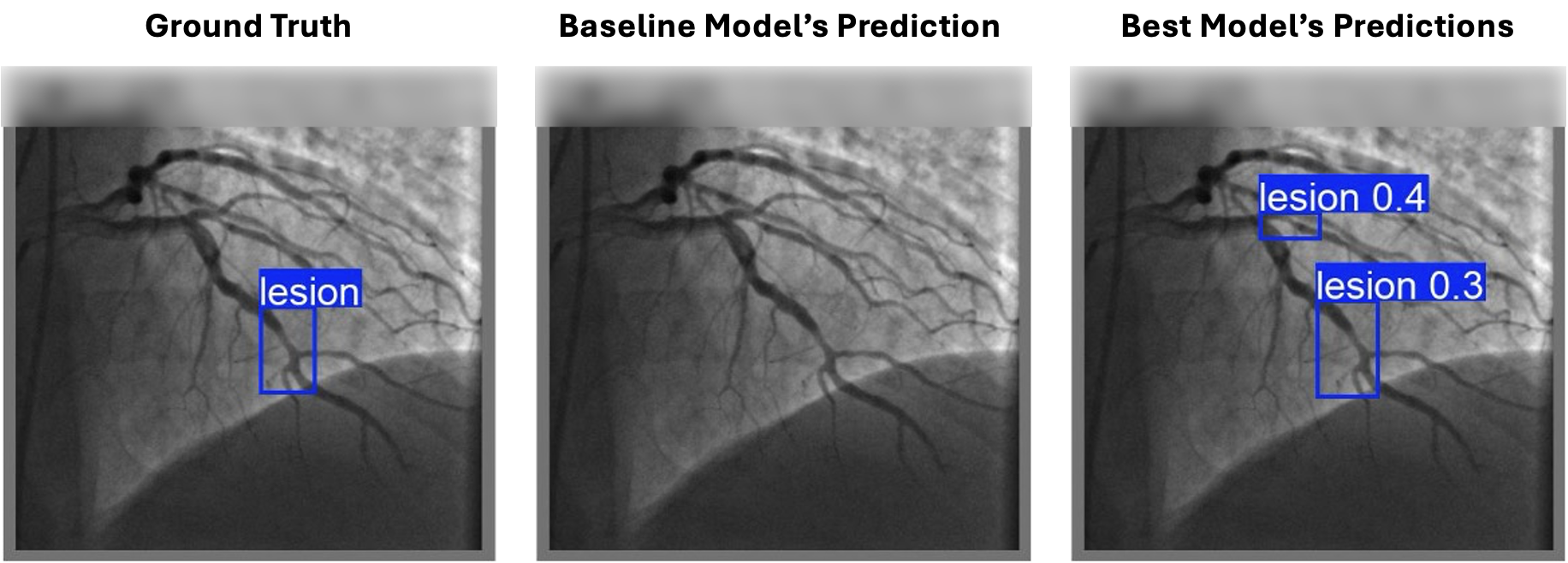}
        }
        \subfigure[]{
            \label{figfvd}
            \includegraphics[scale = 0.12]{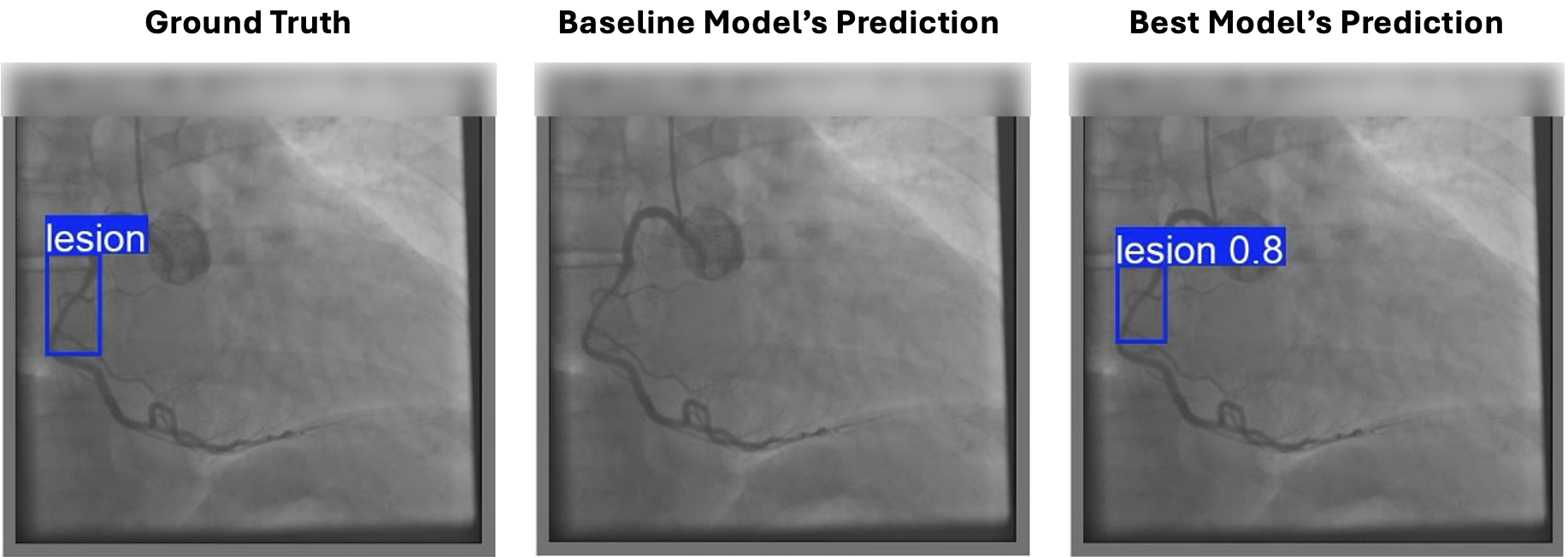}
        }
    \end{center}
    \caption{Lesion Detection using ODySSeI: Visualization of predictions and the corresponding ground truths on a representative sample of the FAME2 validation set show the improved robustness of our best performing lesion detection model over the baseline.}
    \label{figfv}
\end{figure*}

A visualization of our best model's predictions on a representative sample of the validation set alongside those made by the baseline model (see~\autoref{figfv}) clearly demonstrates the benefits of using our PAS. Our best model's predictions are mostly in agreement with the ground truths. Moreover, our best model is also able to detect lesions which are missing from the corresponding ground truth (may not be false positives, see~\autoref{figfvc}, for instance). This is in stark contrast to the baseline model's predictions: the baseline model identifies fewer lesions than our best model (see~\autoref{figfvd} for an illustrative example) and also detects lesions in anatomically unlikely locations in ICA images (see~\autoref{figfva} for e.g.). All in all, these observations highlight the improved robustness of our best performing lesion detection model over the baseline.

\begin{figure*}[!htb]
    \begin{center}
        \subfigure[]{
            \label{figfta}
            \includegraphics[scale = 0.12]{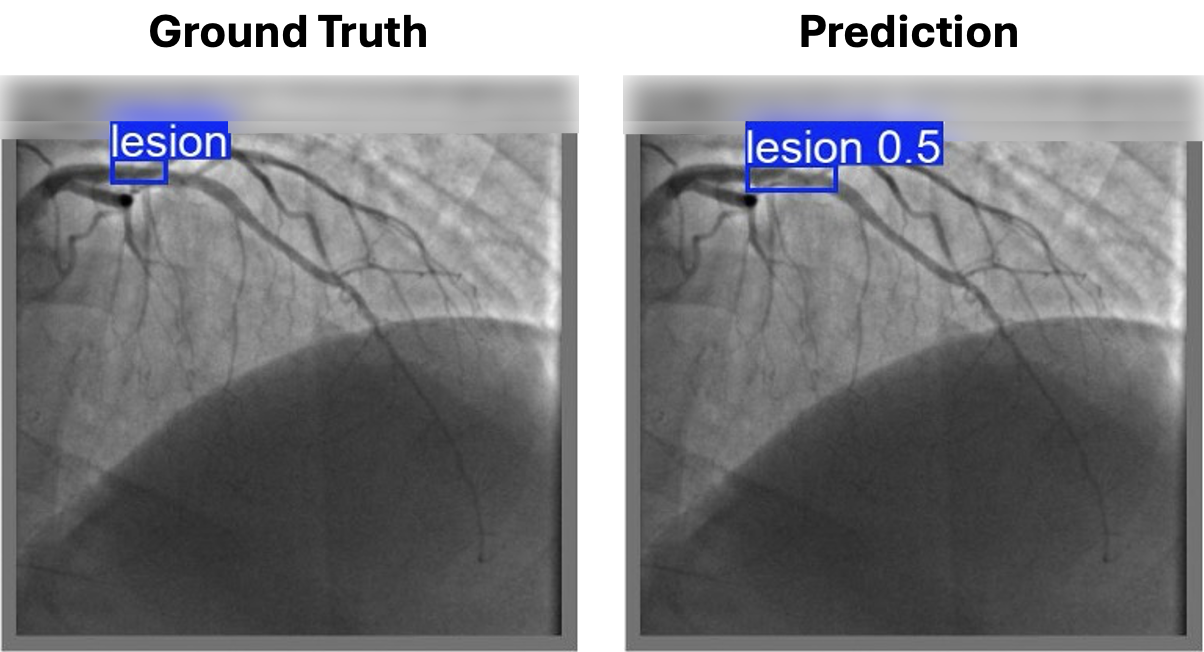}
        }
        \subfigure[]{
            \label{figftb}
            \includegraphics[scale = 0.12]{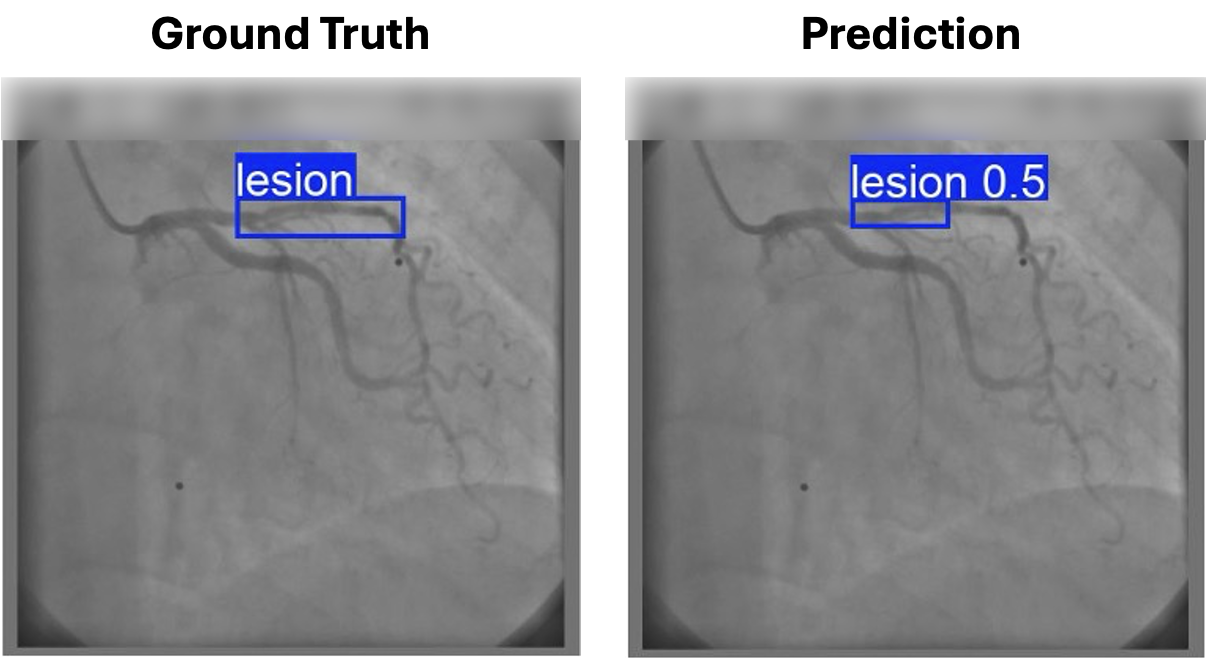}
        }
        \subfigure[]{
            \label{figftd}
            \includegraphics[scale = 0.12]{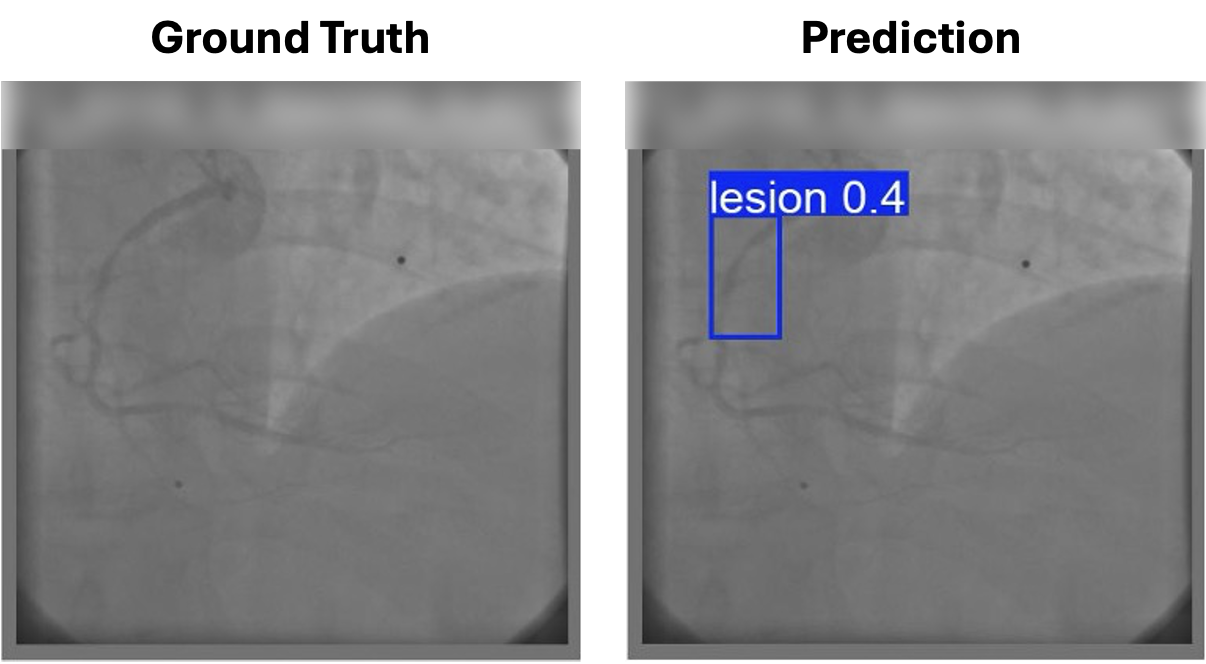}
        }
    \end{center}
    \caption{Lesion Detection using ODySSeI: Visualization of predictions and the corresponding ground truths on a representative sample of the FAME2 test set show that our lesion detection model identifies and localizes lesions accurately and in some cases, identifies lesions which are missing from the corresponding ground truths.}
    \label{figft}
\end{figure*}

\paragraph{\textbf{Test Set Performance}}
\label{final_ldet_res}
\autoref{final_test_set_metrics} shows our best model's performance on the In-Distribution (ID) test set of the FAME2 dataset and the OOD FC dataset. For FAME2, the test set image-level and lesion-level metrics are quite similar and only slightly lower than the corresponding metrics on the validation set (see the last row in~\autoref{fameXarcade}).~\autoref{figft} presents qualitative results from our best performing model on a representative sample of the FAME2 test set. Overall, our model detects lesions quite accurately and, in some cases, it even identifies lesions that are absent from the corresponding ground truth annotations (see~\autoref{figftd} for e.g.). Apart from candidate true positives (false positives which most likely are true lesions), another factor which contributes to lower test set performance is the inconsistency in lesion annotation lengths in the FAME2 dataset. For instance, on examining~\autoref{figfta} and~\autoref{figftb}, we find that the ground truth in~\autoref{figfta} is highly localized while that in~\autoref{figftb} is more widespread. On the other hand, our model's predictions for the corresponding images (see~\autoref{figfta} and~\autoref{figftb}) are more consistent in length and may better reflect the actual lengths of the lesions.

\begin{table}[htbp]
\centering
\caption{Lesion Detection using ODySSeI: Test set performance.}
\resizebox{\textwidth}{!}{%
    \begin{tabular}{|c|c|c|c|c|c|c|c|c|}
        \hline
        \textbf{}&\multicolumn{8}{|c|}{\textbf{Test Set Metrics}}\\
        \cline{2-9}
        \textbf{Dataset}&\multicolumn{4}{|c|}{\textbf{Image-Level$^a$}}&\multicolumn{4}{|c|}{\textbf{Lesion-Level$^b$}}\\
        \cline{2-9}
        \textbf{}&\textbf{Prec$^c$}&\textbf{Rec}&\textbf{mAP@0.50$^c$}&\textbf{mAP@0.50-0.95$^c$}&\textbf{Prec}&\textbf{Rec}&\textbf{mAP@0.50}&\textbf{mAP@0.50-0.95}\\
        \hline
        \textbf{FAME2 (Test)}&\text{$0.578 \pm 0.356$}&\text{$0.839 \pm 0.321$}&\text{$0.646 \pm 0.348$}&\text{$0.337 \pm 0.255$}&0.594&\text{$0.373$}&0.406&0.193\\
        \hline
        \textbf{FC}&\text{$0.338 \pm 0.336$}&\text{$0.672 \pm 0.416$}&\text{$0.492 \pm 0.352$}&\text{$0.226 \pm 0.204$}&
        0.211&0.216&0.132&0.050\\
        \hline
        \multicolumn{6}{l}{\footnotesize{$^a$Mean $\pm$ standard deviation are reported over per-image metrics;}}\\
        \multicolumn{6}{l}{\footnotesize{$^b$Total values are reported (all instances of lesions across all images are considered together);}}\\
        \multicolumn{6}{l}{\footnotesize{$^c$Mean $\pm$ standard deviation are only reported over those images for which the model detects }}\\
        \multicolumn{6}{l}{\footnotesize{ at least one lesion since precision is undefined otherwise. Further, standard deviation values }}\\ 
        \multicolumn{6}{l}{\footnotesize{ are expected to be high since the model can predict FNs. Accordingly, lesion-level metrics, }}\\
        \multicolumn{6}{l}{\footnotesize{ which account for FPs and FNs in total, provide more accurate results.}}\\
    \end{tabular}%
    }
    \label{final_test_set_metrics}
\end{table}

\begin{figure*}[!htb]
    \begin{center}
        \subfigure[]{
            \label{figFCa}
            \includegraphics[scale = 0.12]{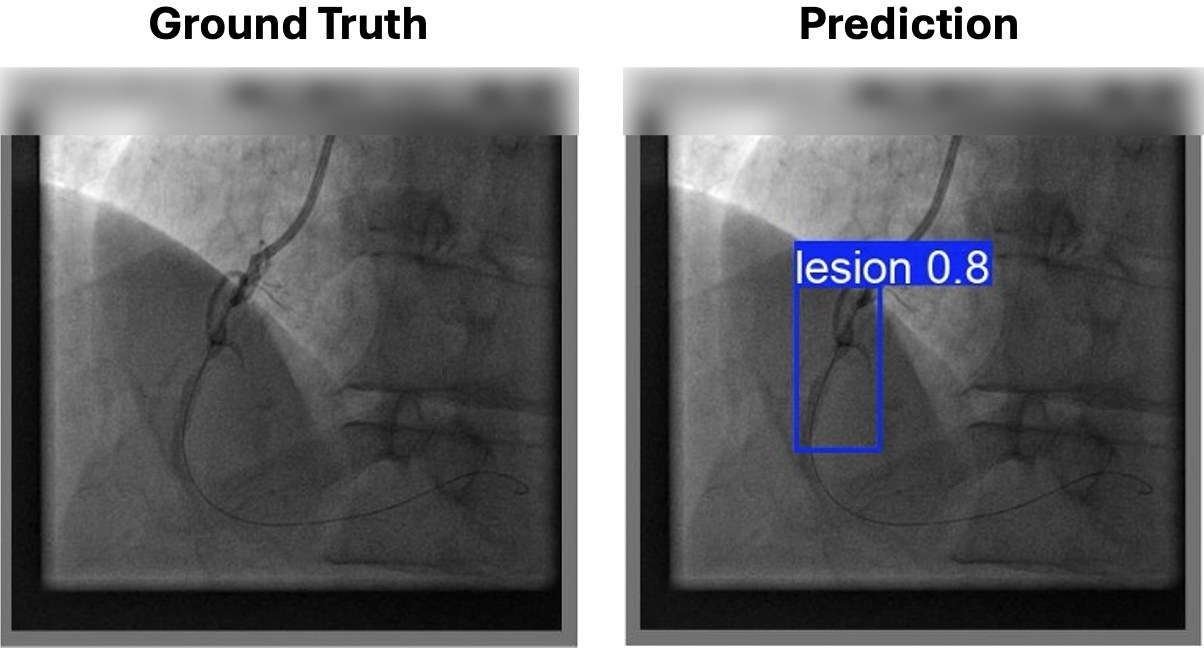}
        }
        \subfigure[]{
            \label{figFCb}
            \includegraphics[scale = 0.12]{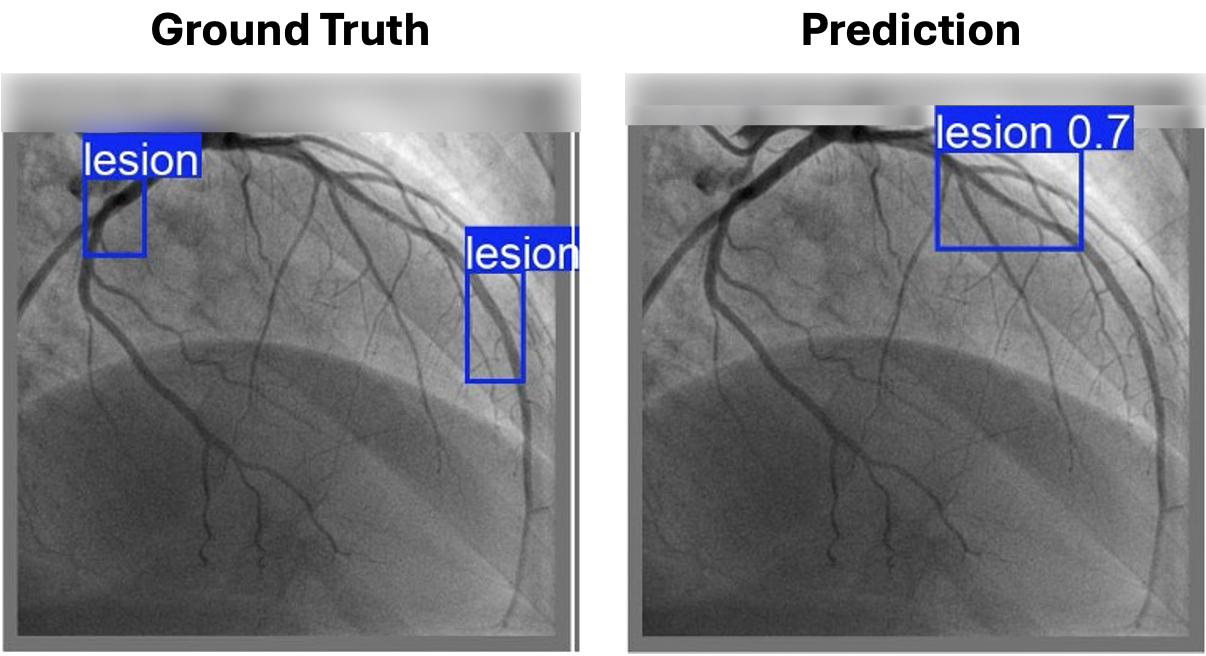}
        }
        \subfigure[]{
            \label{figFCd}
            \includegraphics[scale = 0.12]{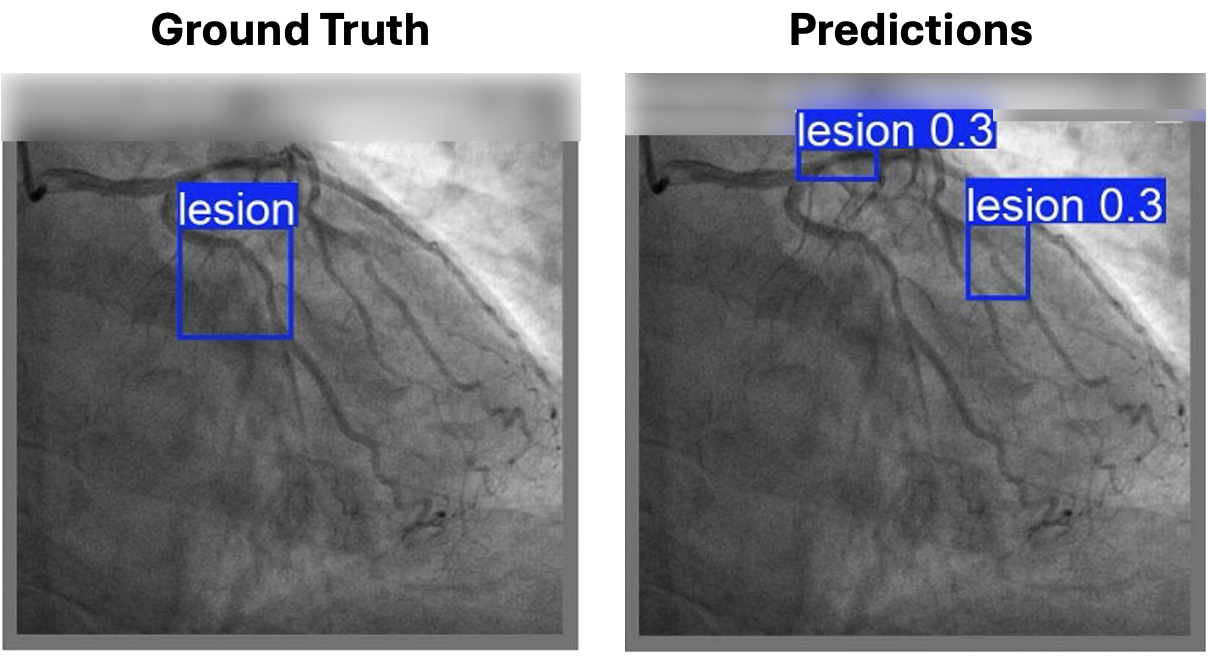}
        }
    \end{center}
    \caption{Lesion Detection using ODySSeI: Visualization of predictions and the corresponding ground truths on a representative sample of the FC dataset show that our model identifies and precisely localizes most of the narrow lesions while occasionally missing out on identifying some wide lesions.}
    \label{figFCt}
\end{figure*}

For the FC dataset, the evaluation metrics are considerably lower than those in the FAME2 test set (see~\autoref{final_test_set_metrics}). This can be qualitatively explained by examining the visualizations of our model's predictions. As shown in~\autoref{figFCt}, while our model occasionally fails to correctly identify wide lesions (possibly due to the inadequate representation of these lesions in the training set), our model identifies most of the lesions where the narrowing is moderate or even high. This is clinically relevant, as such lesions are typically considered the most critical in practice~\cite{FAME2_new}. Furthermore, our model again identifies lesions that are absent from the corresponding ground truth annotations (see~\autoref{figFCa},~\autoref{figFCb}, and~\autoref{figFCd}). In addition, its predictions are more precisely localized than the ground truth annotations, which again remain inconsistent across examples.

\subsubsection{MLD-Based Evaluation}
\label{mld_based_ldet_res}
The performance of ODySSeI's lesion detection model has until now (\autoref{ldet_res}) been evaluated with the help of conventional, bounding box overlap-based metrics. However, as discussed in~\autoref{final_ldet_res}, inconsistencies in lesion annotation lengths and inter-observer variability in the interpretation of ICA images lead to different bounding box coordinates, even for correctly identified lesions. Such variability artificially lowers quantitative scores despite the clinical acceptability of the predictions. To mitigate this subjectivity, we introduce a new set of localization-based metrics which determine whether the bounding box predicted for a given lesion contains the ground truth MLD — the point of maximal stenosis that most directly informs clinical decision-making. The three new metrics: MLD-Precision, MLD-Recall, and MLD-F1 Score, follow the standard definitions of precision, recall, and F1-score, but are computed based on the containment of the exact location of the ground truth MLD (see~\autoref{sec:eval_metrics} for further details).

The results of the MLD-based performance evaluation are shown in~\autoref{mld_metrics}. Our model achieves substantially higher values under these objective metrics than under the bounding box overlap-based metrics shown earlier in Tables~\ref{fameXarcade} and~\ref{final_test_set_metrics}, indicating that the lesion bounding boxes predicted by our model reliably capture the location of the MLD.

\begin{table}[htbp]
\centering
\caption{High values of MLD-based metrics support ODySSeI's clinical reliability. Treating CTPs as (actual) TPs leads to a further boost in ODySSeI's lesion detection performance.}
\resizebox{\textwidth}{!}{%
\begin{tabular}{|c|c|c|c|c|c|c|}
\hline
\text{}&\multicolumn{6}{|c|}{\textbf{MLD-Based Metrics}}\\
\cline{2-7}
\textbf{Dataset}&\multicolumn{3}{|c|}{\textbf{Treating CTPs as FPs}}&\multicolumn{3}{|c|}{\textbf{Treating CTPs as TPs}}\\
\cline{2-7}
\text{}&\textbf{MLD-Precision}&\textbf{MLD-Recall}&\textbf{MLD-F1 Score}&\textbf{MLD-Precision}&\textbf{MLD-Recall}&\textbf{MLD-F1 Score}\\
\hline
\text{FAME2 (Validation)}&0.764&0.545&0.636&0.982&0.607&0.750\\
\hline
\text{FAME2 (Test)}&0.593&0.427&0.496&1.000&0.557&0.715\\
\hline
\text{FC}&0.542&0.198&0.290&1.000&0.313&0.476\\
\hline
\end{tabular}%
}
\label{mld_metrics}
\end{table}

Although these new metrics better reflect clinical relevance, some FPs may turn out to be \emph{candidate} TPs (CTPs), i.e., potentially true lesions which are absent from manual annotations. To systematically evaluate such cases, we employ the non-parametric Mann–Whitney U test~\cite{mann_whitney_u_test} to determine whether the MLD of an FP is a CTP, i.e., it lies in the distribution formed by all the actual ground truth MLDs present in that dataset.

On applying the Mann Whitney U-Test to the MLDs of the FPs for the FAME2 validation and test sets and the FC dataset, we find that 92.308\% (12 out of 13 FPs) of the FPs in the FAME2 validation set, 100\% (24 out of 24 FPs) of the FPs in the FAME2 test set, and 100\% (65 out of 65 FPs) of the FPs in the FC dataset are CTPs. This potentially implies that our model can accurately identify even those lesions which may sometimes be missed by human experts. Upon treating these CTPs as actual TPs, we obtain a further increase in the values of the MLD-based metrics as shown in~\autoref{mld_metrics}. Note that the low values of the MLD-Recall are due to the FNs which are likely due to the variability in lesion identifications amongst cardiologists, a bias inherent to the interpretation of ICA. Hence, MLD-Recall can vary and can potentially even improve with a different set of ground truth annotations. Overall, the high performance of our lesion detection model on our new MLD-based metrics supports its clinical reliability.

\subsection{Lesion Segmentation Performance}
\label{lseg_res}
\subsubsection{Validation Set Performance}

\begin{table*}[htbp]
\centering
\caption{\textbf{PAS-based data augmentation boosts ODySSeI's lesion segmentation performance}. The static augmentation tier consists of seven domain-specific augmentations and the dynamic augmentation tier consists of five probabilistic configurational augmentations. Note that the `none' augmentation suite refers to the baseline model which is trained without any data augmentations.}
\resizebox{\textwidth}{!}{%
\begin{tabular}{|c|c|c|c|c|c|c|c|}
\hline
\multirow{2}{*}{\textbf{Augmentation Suite}}&\multicolumn{7}{|c|}{\textbf{Validation Set Metrics}} \\
\cline{2-8}
\textbf{}&\textbf{Acc}&\textbf{Prec}&\textbf{Rec}&\textbf{Dice}&\textbf{IoU}&\textbf{clDice}&\textbf{MHD}\\
\hline
\textbf{None}&\text{$0.947 \pm 0.033$}&\text{$0.860 \pm 0.094$}&\text{$0.912 \pm 0.074$}&\text{$0.883 \pm 0.076$}&\text{$0.798 \pm 0.106$}&\text{$0.934 \pm 0.081$}&\text{$1.542 \pm 3.351$}\\	
\hline
\textbf{Static}&\text{$0.949 \pm 0.031$}&\text{$0.867 \pm 0.078$}&\text{$0.916 \pm 0.054$}&\text{$0.889 \pm 0.058$}&\text{$0.805 \pm 0.085$}&\text{$0.939 \pm 0.069$}&\text{$1.291 \pm 2.670$}\\
\hline
\textbf{Static + Dynamic}&\text{$\mathbf{0.952 \pm 0.028}$}&\text{$\mathbf{0.876 \pm 0.068}$}&\text{$\mathbf{0.921 \pm 0.049}$}&\text{$\mathbf{0.896 \pm 0.051}$}&\text{$\mathbf{0.816 \pm 0.073}$}&\text{$\mathbf{0.948 \pm 0.060}$}&\text{$\mathbf{0.740 \pm 1.743}$}\\	
\hline
\end{tabular}%
}
\label{full_cropped_seg}
\end{table*}

\begin{figure*}[!htb]
    \begin{center}
        \subfigure[]{
            \label{fig8a}
            \includegraphics[scale = 0.35]{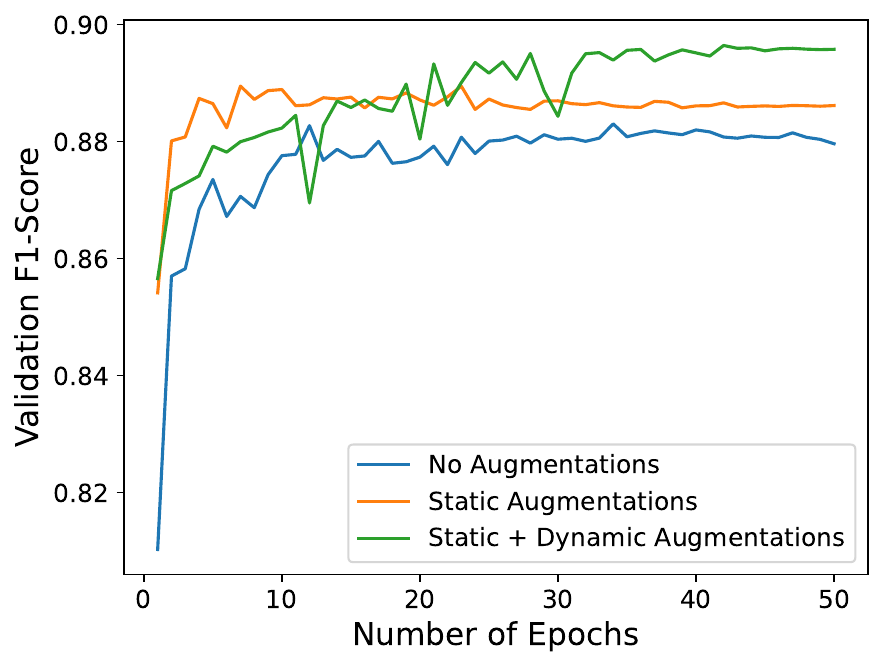}
        }
        \subfigure[]{
            \label{fig8b}
            \includegraphics[scale = 0.35]{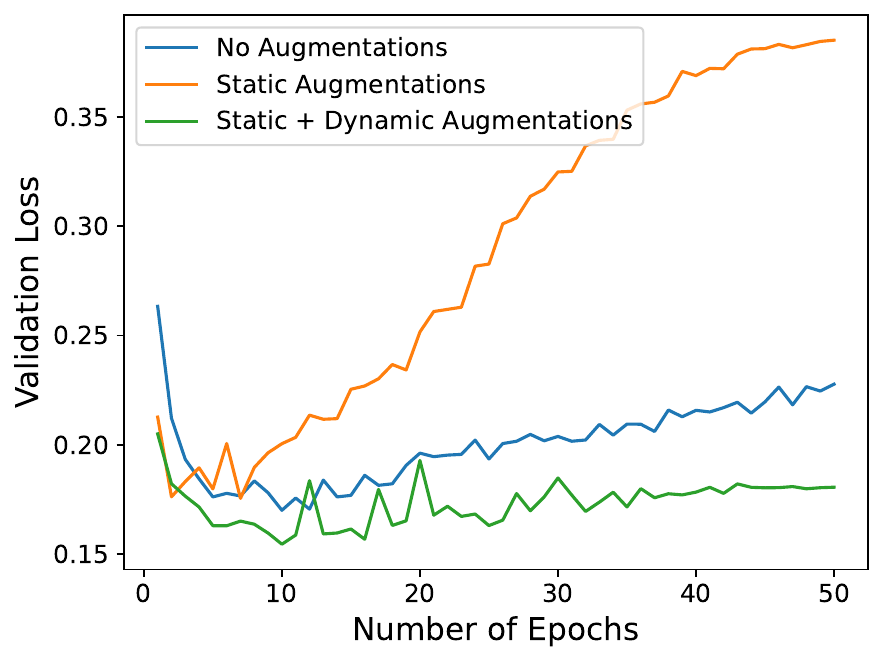}
        }
    \end{center}
    \caption{\textbf{PAS-based data augmentation changes the learning dynamics of ODySSeI's lesion segmentation model} and encourages structured overfitting via static augmentations while imparting robustness and stabilization via dynamic augmentations. Plots show (a) F1-Score on the FAME2 validation set and (b) loss on the FAME2 validation set.}
    \label{fig7}
\end{figure*}

\autoref{full_cropped_seg} shows the effect of using our PAS on the performance of our lesion segmentation model, which is based on the DeepLabv3+ architecture~\cite{deeplabv3+}. Clearly, our PAS enhances the performance of the lesion segmentation model on the cropped FAME2 validation set with 1-3\% improvements across all metrics over the baseline (the model trained on the non-augmented, cropped FAME2 training set).

From~\autoref{fig7}, we again observe why our PAS works well: static augmentations encourage overfitting to diverse real-life ICA scenarios, i.e., \emph{structured} overfitting, while dynamic augmentations provide robustness and stability across plausible configurational changes in image acquisition. Notably, our PAS leads to a larger performance boost over the baseline when the task is more complex (2.5-fold in lesion detection versus 1-3\% in lesion segmentation). This is expected because learning from diverse real-world augmentations benefits the most when the task is complex and ambiguous, as is the case for lesion detection, since our PAS provides denser supervision by exposing the model to multiple plausible variants of only a sparse set of lesion instances. In contrast, the relatively simpler task of lesion segmentation already provides dense supervision (pixel-level labels) and is typically less ambiguous, so, performance improvements tend to be incremental once the model captures the core structure of the task.

\subsubsection{Test Set Performance}
\begin{figure*}[htbp]
    \begin{center}
        \subfigure[]{
            \label{figdl1}
            \includegraphics[scale = 0.25]{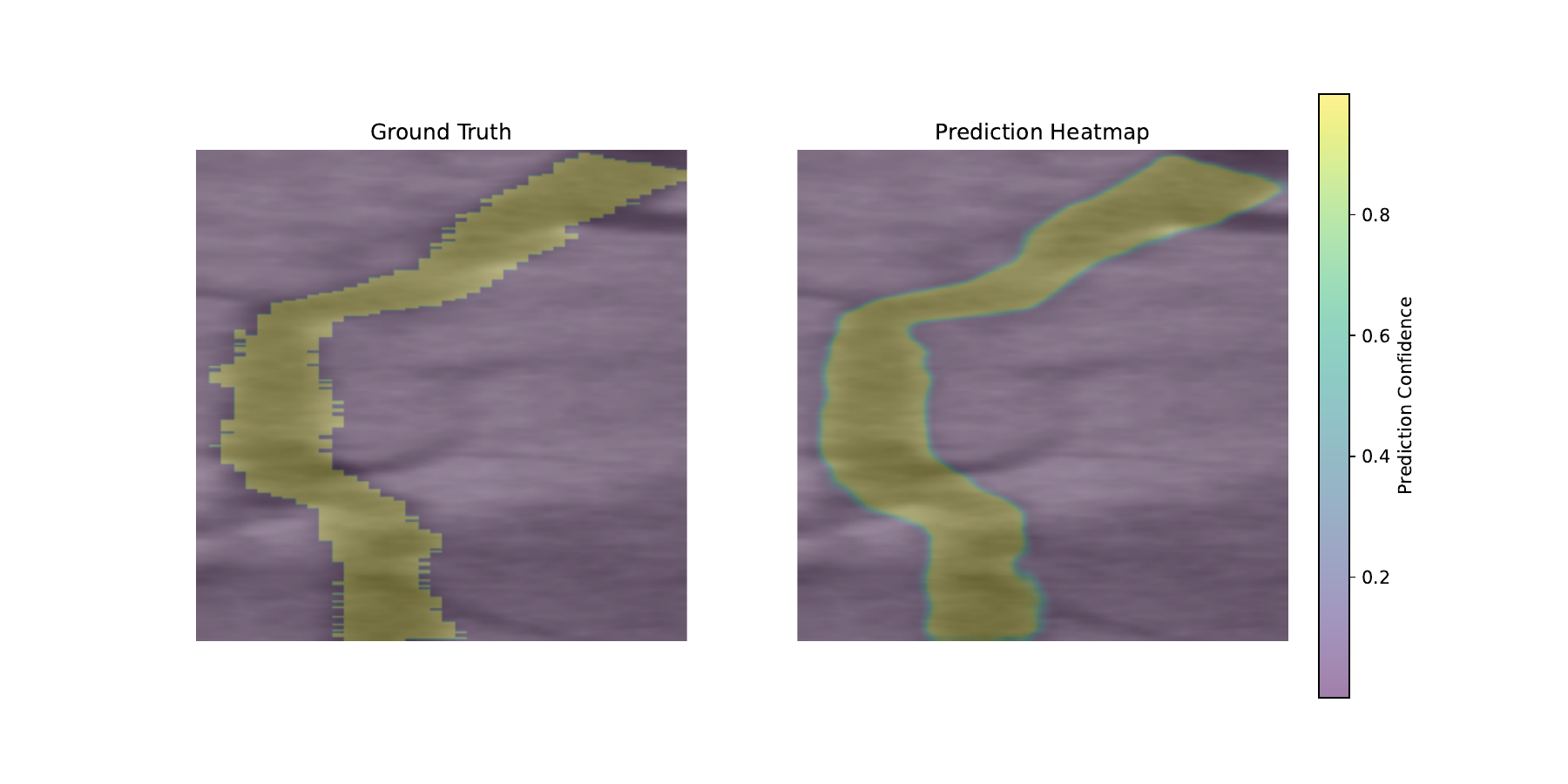}
        }
        \subfigure[]{
            \label{figdlFC2}
            \includegraphics[scale = 0.25]{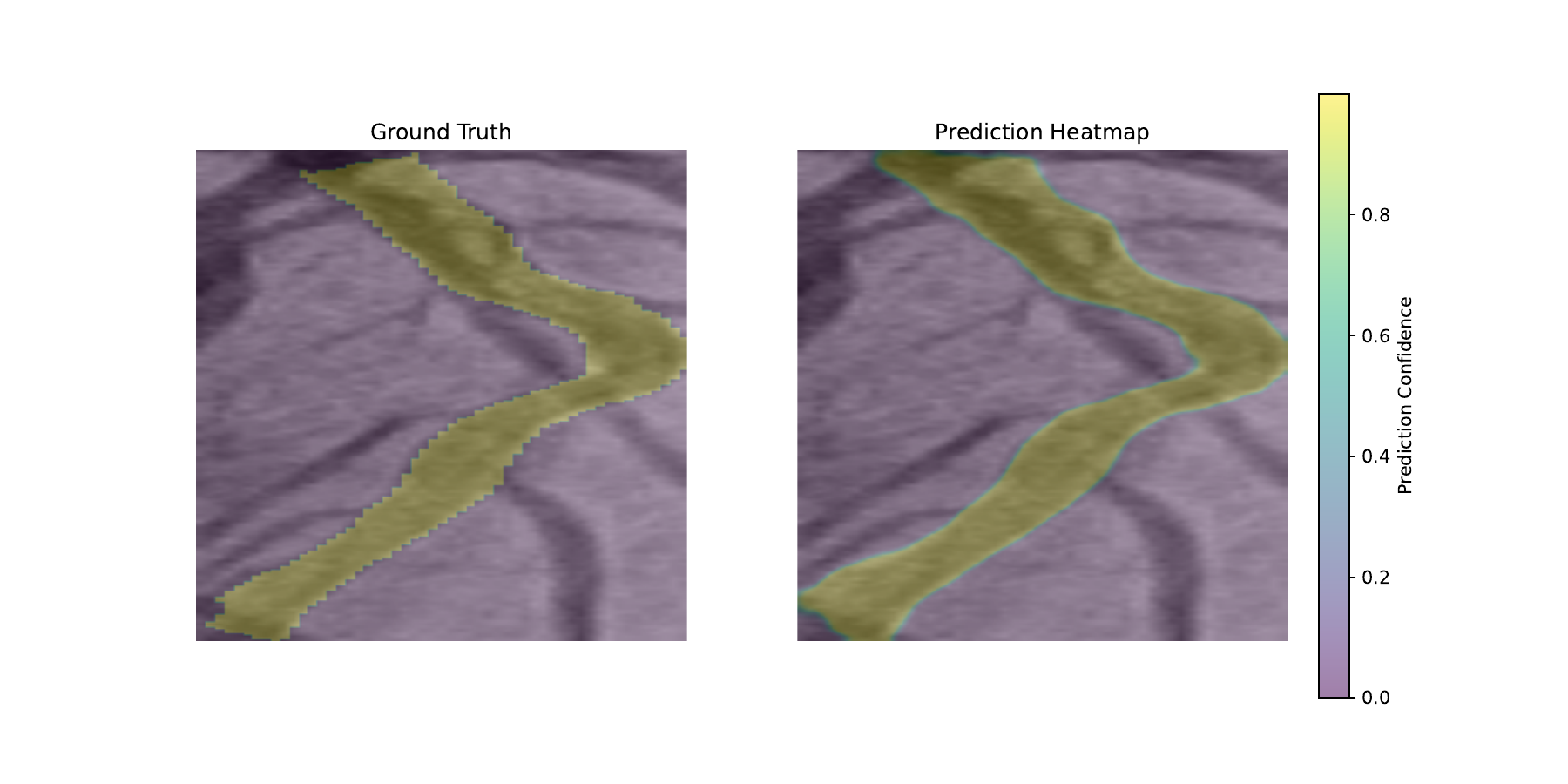}
        }
    \end{center}
    \caption{Lesion Segmentation using ODySSeI: Visualization of predictions and the corresponding ground truths on a representative sample of (a) the cropped FAME2 test set and (b) the cropped FC dataset show that our model segments lesions well across different cohorts.}
    \label{figdl_lseg}
\end{figure*}

\autoref{deeplab_fame_test} and~\autoref{figdl_lseg} show that our model accurately segments lesions with precision, by respecting their topology, and ignoring any overlapping arterial structures. Further, from~\autoref{figdl_lseg}, it is clear that our model's lesion segmentations are better than the ground truths: the former are smooth at the lesion boundaries while the latter are extremely rough at the edges. Interestingly, our best lesion segmentation model performs better on the OOD dataset than the ID test set (exactly the opposite of what was seen in lesion detection; see~\autoref{final_test_set_metrics}), likely due to wider lesions in the OOD dataset which are easier to segment than narrow lesions in the ID test set due to higher spatial contiguity of lesion pixels in the former than the latter. 
\begin{table*}[htbp]
\centering
\caption{Lesion Segmentation: Test set performance.}
\begin{tabular}{|c|c|c|c|c|c|c|c|}
\hline
\textbf{}&\multicolumn{7}{|c|}{\textbf{Test Set Metrics}} \\
\cline{2-8}
\textbf{Dataset}&\textbf{Acc}&\textbf{Prec}&\textbf{Rec}&\textbf{Dice}&\textbf{IoU}&\textbf{clDice}&\textbf{MHD}\\
\hline
\text{FAME2 (Test)}&\text{$0.941 \pm 0.040$}&\text{$0.867 \pm 0.065$}&\text{$0.935 \pm 0.031$}&\text{$0.898 \pm 0.038$}&\text{$0.817 \pm 0.060$}&\text{$0.948 \pm 0.048$}&\text{$0.796 \pm 1.309$}\\		
\hline
\text{FC}&\text{$0.958 \pm 0.034$}&\text{$0.913 \pm 0.052$}&\text{$0.946 \pm 0.061$}&\text{$0.929 \pm 0.052$}&\text{$0.870 \pm 0.075$}&\text{$0.963 \pm 0.057$}&\text{$0.713 \pm 2.343$}\\
\hline
\end{tabular}
\label{deeplab_fame_test}
\end{table*}

\subsection{Lesion Severity Estimation}
\label{lse_res}

\begin{figure*}[!htb]
    \begin{center}
        \subfigure[]{
            \label{figba1}
            \includegraphics[scale = 0.33]{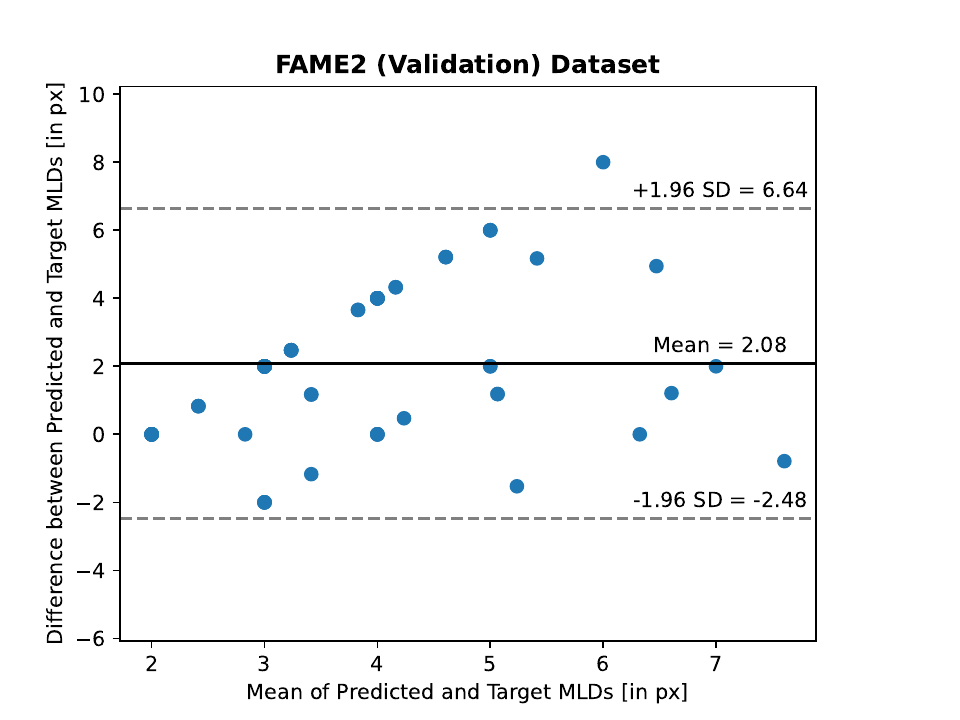}
        }
        \subfigure[]{
            \label{figba2}
            \includegraphics[scale = 0.33]{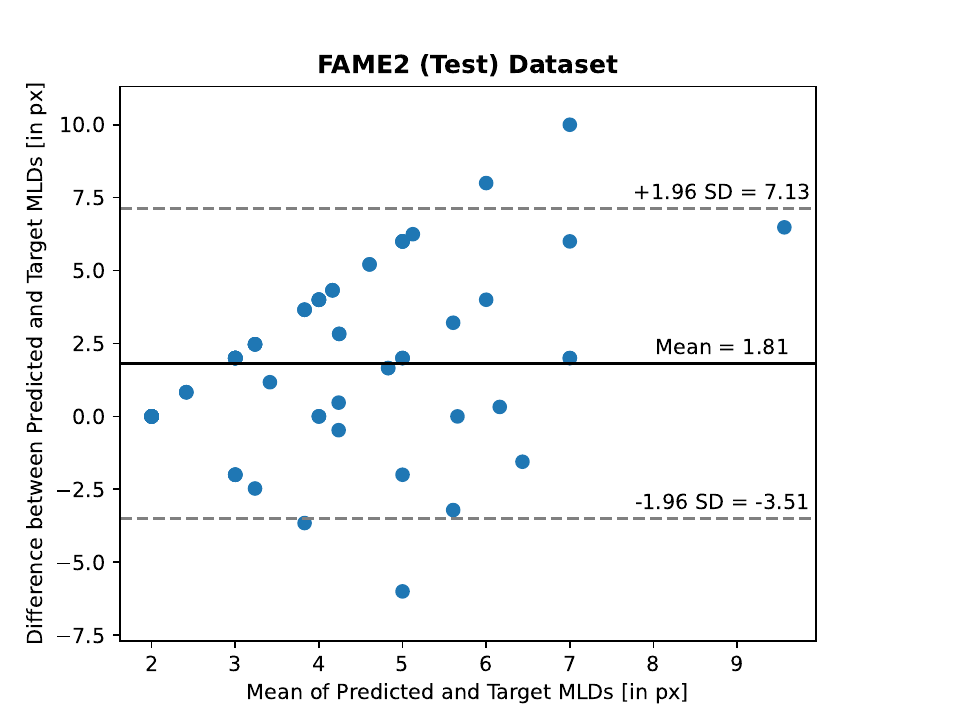}
        }
        \subfigure[]{
            \label{figba3}
            \includegraphics[scale = 0.33]{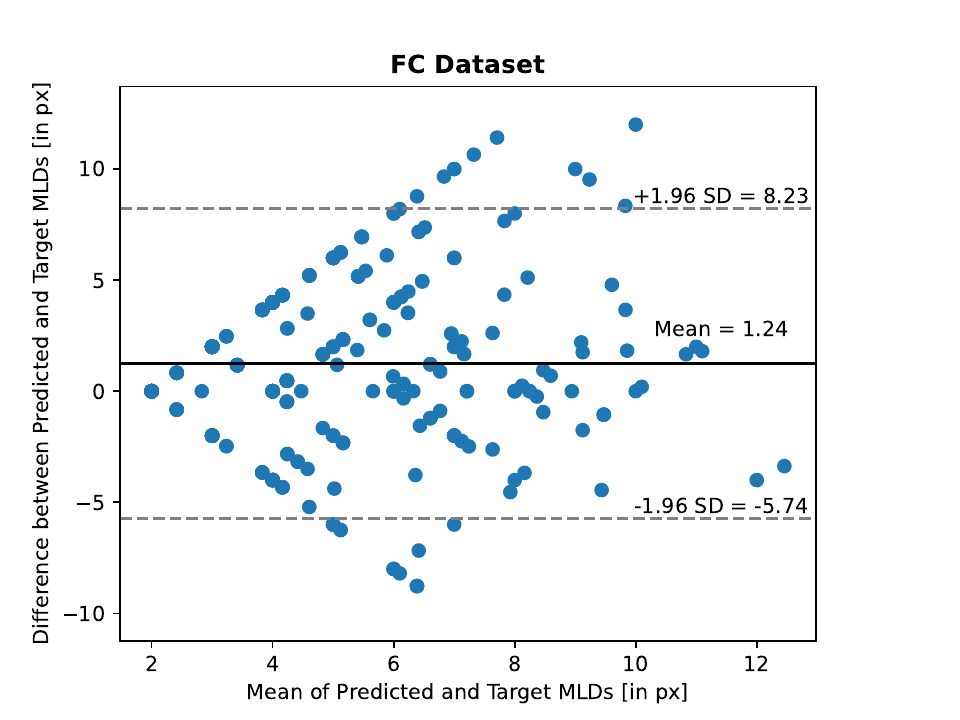}
        }
    \end{center}
    \caption{{Lesion Severity Estimation: Bland-Altman plots show high agreement between predicted and ground truth MLDs} for the (a) FAME2 validation set, the (b) FAME2 test set, and the (c) FC dataset.}
    \label{figba}
\end{figure*}

High F1 scores and balanced accuracies (see~\autoref{final_LSE_metrics}) together with low mean differences (see~\autoref{figba}) between predicted and ground truth MLDs, across both ID as well as OOD datasets, show that the MLDs computed via the usage of our lesion severity estimation algorithm are in high agreement with the corresponding ground truths. The reason behind the $\pm$ 2-3 pixel (px) Mean Absolute Difference (MAD) also follows from~\autoref{figdl_lseg}: smooth lesion boundaries predicted by our model are bound to deviate from rough boundaries in the corresponding ground truths by a few px.

\begin{table}[htbp]
\centering
\caption{MLDs predicted with our QCA-free lesion severity estimation technique are in high agreement with the corresponding ground truths for both ID as well as OOD datasets.}
\begin{tabular}{|c|c|c|c|c|c|}
\hline
\textbf{Dataset}&\multicolumn{5}{|c|}{\textbf{Metrics}}\\
\cline{2-6}
\text{}&\text{MAD$^a$ $\pm$ SD$^b$}&\text{Prec$^d$ (95\% CI$^e$)}&\text{Rec$^d$ (95\% CI$^e$)}&\text{F1$^d$ (95\% CI$^e$)}&\text{Bal Acc$^d$ (95\% CI$^e$)}\\
\hline
\text{FAME2 (Validation)}&2.378 $\pm$ 2.023&1.000 (0.657-1.000)&0.656 (0.400-0.664)&0.792 (0.505-0.794)&0.828 (0.363-0.835)\\
\hline
\text{FAME2 (Test)}&2.470 $\pm$ 2.133&1.000 (0.643-1.000)&0.656 (0.394-0.659)&0.792 (0.495-0.798)&0.828 (0.366-0.836)\\
\hline
\text{FC}&2.877 $\pm$ 2.444&1.000 (0.567-1.000)&0.684 (0.378-0.700)&0.812 (0.466-0.877)&0.842 (0.449-0.851)\\
\hline
\multicolumn{6}{l}{\footnotesize{$^a$Mean Absolute Difference; $^b$Standard Deviation; $^c$Prec, Rec, F1, and Bal Acc are used to assess our}}\\
\multicolumn{6}{l}{\footnotesize{algorithm’s performance in determining lesions with an MLD $\leq$ 4 px. Note that a threshold of 6 px is used}}\\
\multicolumn{6}{l}{\footnotesize{to binarize our algorithm’s predictions to account for the $\pm$ 2 px difference in predicted and ground truth }}\\
\multicolumn{6}{l}{\footnotesize{MLDs which accrue due to smooth (predicted) and rough (ground truth) lesion boundaries as discussed;}}\\ 
\multicolumn{6}{l}{\footnotesize{$^d$ 95\% CIs are determined by bootstrapping the data over 1000 iterations.}}\\
\end{tabular}
\label{final_LSE_metrics}
\end{table}

\section{Discussion}
In this work, we introduce ODySSeI, an end-to-end framework for automated lesion detection, lesion segmentation, and lesion severity estimation in ICA images. Our proposed PAS with internal augmentation mixing enables the generation of large-scale realistic training data, facilitating improved lesion detection and segmentation. This is in contrast to prior works~\cite{data_aug_survey} which neither combine augmentation tiers nor increase the data volume with them. 

Our PAS incorporates clinically relevant imaging variations, including changes in radiographic contrast as well as degradations caused by noise, occlusion, and out-of-focus imaging during ICA. It also integrates domain-informed transformations that reflect stenosed and normal arterial geometry and topology, thereby improving robustness to real-world clinical variability. This is exemplified by our PAS' utility in consistently enhancing model performance with large gains in highly complex tasks as compared to relatively simpler ones (2.5-fold in lesion detection versus 1-3\% in lesion segmentation). Furthermore, our experimentation with different tiers of our PAS brings to light their characteristic functionalities: static augmentations enforce structured overfitting by promoting the learning of diverse real-life scenarios; dynamic augmentations make our models more robust to plausible configurational changes, increase and stabilize their generalizability; composite augmentations help in learning rich compositionality and increase the precision in lesion localization.

Our QCA-free lesion severity estimation technique leverages the stenosed and normal arterial geometry and topology for a rule-based computation of MLD and DS that negates reliance on any training data. Low MADs of just $\pm$ 2 - 3 px with respect to the actual ground truth MLDs confirm its high accuracy. Prior work trained AI models for predicting severity metrics which led to high MADs on their test data owing to their high dependence on the training data~\cite{CathAI,DeepCoro}. As compared to lesion detection and lesion segmentation which cannot possibly be automated via a simple rule-based approach (and hence, require the usage of complex AI models) due to the diverse ways in which lesions manifest in ICA images, computing MLD and DS from lesion segmentation maps simply relies on determining the amount of minimal and maximal arterial narrowing which are geometry and topology dependent only and hence, can be efficiently automated via a rule-based approach.

Given the subjective nature of ICA analysis which lends it to high intra- and inter-operator variability~\cite{ICA_variability}, bounding box overlap-based performance metrics, which are highly sensitive to changes in bounding box coordinates, are bound to differ on comparing our lesion detection model's predictions against another set of expert annotations for the same set of ICA images. Determining the concordance of predictions with another set of expert annotations, as done by prior work~\cite{CathAI,DeepCoro}, can only partially alleviate this issue of subjectiveness in performance evaluations since they might still turn out to be less concordant, i.e., more discordant, with yet another set of expert annotations. To the best of our knowledge, our work is the first to develop a more objective set of metrics based on the confinement of the exact location of the MLD within the predicted bounding boxes for lesions (if any) in ICA images. The location of the MLD is central to clinical decision-making and does not vary for a given lesion from one expert annotator to another. High performance under these metrics indicates that our model reliably captures the anatomically meaningful region of maximal stenosis.

Two recent end-to-end ICA analysis frameworks are CathAI~\cite{CathAI} and DeepCoro~\cite{DeepCoro}. As compared to these frameworks, ODySSeI adopts a substantially simpler architecture, requiring only two deep learning models instead of five~\cite{CathAI} and twelve deep learning models~\cite{DeepCoro}, respectively. This design leads to its improved memory and compute efficiency, enabling real-time processing with an average input processing time of just a fraction of a second on a GPU and 5 seconds on a CPU. In contrast, CathAI and DeepCoro have substantially longer average processing times (2 seconds~\cite{CathAI} and 62.6 seconds~\cite{DeepCoro} on a GPU, respectively; their CPU processing times are unstated) limiting their suitability for real-time clinical deployment. Furthermore, unlike CathAI and DeepCoro which exclude side branch lesions during training~\cite{CathAI,DeepCoro}, ODySSeI is trained using all the available images without enforcing any post-hoc exclusion criteria, allowing it to detect and segment lesions even in complex coronary side branches. Additionally, while CathAI and DeepCoro heavily underestimate the severity of narrow (often \emph{critical}) lesions (for performance evaluations, CathAI and DeepCoro authors consider lesions with predicted severities $\geq$ 54\% and $\geq$ 23\% (respectively) as possessing actual clinical severities $\geq$ 70\%)~\cite{CathAI,DeepCoro}, ODySSeI's estimates of lesion severity (MLD) just possess a $\pm$ 2-3 px difference w.r.t. the MLDs of the ground truths. Finally, in contrast to CathAI and DeepCoro which are closed-source (only the lesion segmentation models of DeepCoro are open-source~\cite{DeepCoro}) and require images in specific formats with additional metadata as inputs~\cite{CathAI,DeepCoro}, ODySSeI is released as an open-source framework with a plug-and-play web interface, facilitating clinical usability and reproducibility.

Overall, ODySSeI advances end-to-end ICA analysis by combining robust lesion detection and segmentation, objective evaluation aligned with clinical relevance, and accurate lesion severity estimation without QCA or additional training. However, it does have certain limitations. Firstly, ODySSeI still requires a cardiologist to select a high resolution, end-diastolic ICA frame (image) from a patient's ICA video as the input for ICA analysis. Secondly, although ODySSeI is trained with ICA images from multiple sites across the world, the size of our training sets is limited. Future work involves automating end-diastolic ICA frame selection, large-scale expert annotation, and prospective clinical validation studies to test the efficacy of ODySSeI in clinical settings. 

\section{Methods}

\subsection{Datasets}
\label{datasets}

In this study, we use the following three datasets which include two in-house datasets and one public dataset:

\begin{enumerate}
\item \textbf{FAME2}: The \textbf{F}ractional Flow Reserve versus \textbf{A}ngiography for \textbf{M}ultivessel \textbf{E}valuation \textbf{2} dataset is collected via a randomized trial conducted in 28 sites across Europe and North America. It examines the benefits of Percutaneous Coronary Intervention (PCI) as an addition to the best available medical therapy in one-, two-, or three-vessel stable coronary heart disease patients with Fractional Flow Reserve (FFR)-based functionally significant stenoses~\cite{FAME2_orig}. The primary end-point is the Vessel-Oriented Clinical End-Point (VOCE) at 2 years which is a composite of prospectively adjudicated cardiac death, Myocardial Infarction (MI), and urgent and non-urgent revascularization~\cite{FAME2_new}. We utilize the 2D Digital Imaging and Communications in Medicine (DICOM) images obtained via ICA in 566 patients. This amounts to a total of 1735 grayscale (512 $\times$ 512) images, out of which 1352 images have at least 1 lesion. The average age of the patients is 63.9 $\pm$ 9.8 years with 145 (25.6\%) identifying as female and 421 (74.4\%) identifying as male~\cite{FAME2_orig,FAME2_new}. The lesions in these angiograms have been annotated by an expert team from the Department of Cardiology at CHUV, Lausanne, Switzerland. The training, validation, and test splits consist of 1561, 86, and 88 images respectively. 

\item \textbf{ARCADE}: The \textbf{A}utomatic \textbf{R}egion-based \textbf{C}oronary \textbf{A}rtery \textbf{D}isease Diagnostics using X-Ray Angiography Imag\textbf{E}s dataset has been retrospectively obtained from the X-ray and vascular surgery archive of the Research Institute of Cardiology and Internal Diseases, Almaty, Kazakhstan~\cite{ARCADE}. We utilize the training set of the stenosis detection subset which comprises 1000 grayscale (512 $\times$ 512) images, from 1500 patients, containing a single lesion each. The average age of the patients is 45.8 years with 645 (43\%) identifying as female and 855 (57\%) identifying as male~\cite{ARCADE}. LAD, LCX, and RCA arteries have been equally represented in the dataset and hence, provide for a balanced analysis.

\item \textbf{Future Culprit}: This dataset has been obtained via a retrospective multicentre case–control study conducted in three PCI centres in Europe: Lausanne University Hospital, Switzerland; Fribourg University Hospital, Switzerland; OLV Aalst, Belgium~\cite{Future_Culprit}. The study comprises 83 patients who have had an MI and have undergone ICA in the 5 years preceding the MI (baseline ICA). The average age of the patients is 68.8 ± 12.8 years with 23 (27.7\%) identifying as female and 60 (72.3\%) identifying as male~\cite{Future_Culprit}. We utilize the 302 baseline ICA images for this study. 
\end{enumerate}

\subsection{Components of ODySSeI}

\subsubsection{Lesion Detection}
\label{detection}
We use YOLO11m~\cite{yolo11} (pre-trained on the COCO dataset~\cite{coco}) for lesion detection due to its higher localization accuracy and lower latency than competitive models such as Faster R-CNN~\cite{faster_rcnn} and EfficientDet~\cite{efficientdet}. We use a weighted combination of three loss functions for training~\cite{yolo11}: Complete Intersection over Union Loss (7.5$\mathcal{L}_{CIoU}$)~\cite{ciou} for precise localization, Distribution Focal Loss (1.5$\mathcal{L}_{DFL}$)~\cite{dfl} to further enhance detection performance in ambiguous and/or challenging scenarios, and VariFocal Loss (0.5$\mathcal{L}_{VFL}$)~\cite{vfl} to address class imbalance by promoting IoU-aware classification.

\subsubsection{Lesion Segmentation}
\label{segmentation}
We use DeepLabv3+~\cite{deeplabv3+} (pre-trained on the ImageNet-1K dataset~\cite{imagenet}) for lesion segmentation since it learns rich semantic information and performs precise segmentation. We use an equally weighted combination of soft-Dice~\cite{soft_dice} and weighted BCE loss~\cite{unet} to encourage high pixel-wise agreement.

\subsubsection{Lesion Severity Estimation}
\label{severity}
The Minimum Lumen Diameter (MLD) and the percentage Diameter Stenosis (DS) are often used to quantify lesion severity~\cite{FAME2_new}. To reduce the burden of manual intervention, we propose an innovative QCA-free technique (see~\autoref{lse}) for computing these measures directly from lesion segmentation maps.

Firstly, we extract the centerline skeleton~\cite{skeleton} of the lesion from the segmented image. Secondly, for every pixel in the lesion (foreground) skeleton, we compute its distance to the nearest background pixel using the Euclidean distance transform map. Using the distance transform has the distinct advantage of allowing us to correctly compute the radius of an arterial portion since it intrinsically computes the shortest distance which is the distance normal to the radius of curvature at every point. Thirdly, in order to extract the MLD from the computed array of radii, we extract the minimum between two radii peaks and double it. This is motivated by the observation that maximal narrowing typically occurs between two healthy arterial segments; twice of the global minimum radius may reflect boundary artifacts (e.g., tapering near cropped segmentation edges) rather than true stenosis. Finally, the Maximal (Healthy) Arterial Diameter (MAD) can be computed by doubling the radius of the highest peak. The DS can then be computed using the following equation: $$DS = \left(1 - \frac{MLD}{MAD}\right)\times 100\%.$$

\subsection{Pyramidal Augmentation Scheme (PAS)}

In medical imaging, annotated datasets are often limited in size because expert labeling is time-consuming and costly. To address this, we propose a novel and customizable Pyramidal Augmentation Scheme (PAS) which increases the effective training set size and improves robustness to clinically plausible variations during real-world inference (see~\autoref{pas}). The three components of our PAS are as follows:
\begin{enumerate}
    \item \textbf{Static Augmentations}: To learn \emph{global} and \emph{local} arterial geometry and topology, while negating reliance on noise or specific contrast as shortcuts, and to simulate real-world scenarios such as speckle noise and out-of-focus imaging in ICA, we use seven different \emph{domain-specific} augmentations: \textbf{C}ontrast \textbf{L}imited \textbf{A}daptive \textbf{H}istogram \textbf{E}qualization (\textbf{CLAHE})~\cite{CLAHE_1}, Inversion, Multiplicative Noise, Median Blur, Motion Blur, Defocus Blur, and Local Pixel Shuffling~\cite{lps} to make our new training dataset eight times larger than the original one. They do not affect the ground truth and are fixed (static) before training since they are implemented \emph{a priori}.
\begin{enumerate}
        \item CLAHE is a tile-based method for contrast enhancement which performs balanced histogram equalization by preventing overamplification of noise~\cite{CLAHE_1,CLAHE_2}. Inversion inverts the input image by subtracting pixel values from the maximal pixel value while at the same time preserving the original relative pixel intensities, by for instance, transforming black to white and vice versa. Multiplicative Noise simulates the effect of speckle noise which frequently occurs in X-Ray ICA. All these three transforms make our lesion detection and lesion segmentation models focus on the lesion geometry and topology rather than relying on noise or specific contrast as shortcuts for making correct predictions. 
        \item The three different types of blurring enable a holistic simulation of the effects of out-of-focus imaging by the X-Ray camera in ICA. Median Blur performs kernel-wise median-guided replacement and removes salt-and-pepper noise while preserving edges. Motion Blur affects angular and directional blurring while Defocus Blur is aperture-based static disc kernel blurring.  
        \item Local Pixel Shuffling performs shuffling of the image pixels in small windows of the image~\cite{lps} and thus forces the model to learn image texture and focus on global, \textit{contextual} information in the image and not simply rely on local shortcuts.
\end{enumerate}
    \item \textbf{Dynamic Augmentations}: To counteract overfitting and enhance generalization, we probabilistically apply dynamic augmentations to the statically augmented dataset during training. We use five different \emph{general} augmentations: Random Scaling (encourages depthwise focus), Random Erasing (in-painting for adversarial training~\cite{erasing}), Random Translation, Random Color Jiggle (simulates different lighting conditions), and Random Horizontal Flips (encourages learning of symmetry). These are more general augmentations than those in the static tier and do not increase the dataset size. 
    \item \textbf{Composite Augmentations}: To simulate occlusion and enable compositionality through fragmentation, we apply the Mosaic augmentation~\cite{mosaic}, but only until the last 10 epochs, to stabilize training towards the end. This tier is exclusively used for training our lesion detection model since lesion segmentation does not benefit from composite scene understanding given that it is highly localized.
\end{enumerate}

As shown in~\autoref{pas}, our PAS incorporates internal mixing since the upper augmentation tier is always applied on top of the lower augmentation tier. Static augmentations are performed using the Albumentations library~\cite{albumentations} while dynamic augmentations are performed using PyTorch~\cite{pytorch} and Kornia~\cite{kornia} (for efficient batch augmentations).

\subsection{Experimental Settings}

\subsubsection{Evaluation Metrics}
\label{sec:eval_metrics}
\begin{enumerate}
    \item \textbf{Lesion Detection}: We use the following set of bounding box overlap-based metrics for performance evaluation: Precision (Prec), Recall (Rec), mean Average Precision (mAP) over an Intersection over Union (IoU) threshold of 0.50 (mAP@0.50), and mAP over an IoU threshold range of 0.50-0.95 (mAP@0.50-0.95). For MLD-based evaluation, the new metrics proposed by us are as follows:
\begin{align*}
    \text{MLD-Precision} = \left(\dfrac{\text{TP}}{\text{TP} + \text{FP}}\right),\quad \text{MLD-Recall} = \left(\dfrac{\text{TP}}{\text{TP} + \text{FN}}\right)\,,\,\\ \text{MLD-F1 Score} = \left(\dfrac{2\times\text{MLD-Precision}\times\text{MLD-Recall}}{\text{MLD-Precision} + \text{MLD-Recall}}\right),
\end{align*}
where, TPs, FPs, and FNs are defined as follows:
\begin{enumerate}
    \item \textbf{TP}: A lesion bounding box predicted by the model in an ICA image is said to be a TP when it contains an MLD from the corresponding ground truth.
    \item \textbf{FP}: A lesion bounding box predicted by the model in an ICA image is said to be an FP when it does not exist in the ground truth, i.e., no corresponding ground truth MLD is contained within it (if a corresponding ground truth lesion exists for that ICA image) or no corresponding ground truth MLD is present in the first place (no corresponding ground truth lesion exists for that ICA image).
    \item \textbf{FN}: A ground truth lesion bounding box in an ICA image is said to be an FN when either the model fails to detect the corresponding lesion bounding box or when the MLD of the corresponding ground truth lesion is not contained within any of the lesion bounding boxes predicted by the model. 
\end{enumerate}
    Finally, note that the Mann Whitney U-Test is a non-parametric test which measures whether the distributions of two samples are equivalent or not. However, the test-statistic (and its corresponding critical value) by virtue of which it checks the validity of this null hypothesis can serve as a robust method to determine whether the MLD of an FP lies in the distribution formed by all the actual ground truth MLDs present in that dataset~\cite{mann_whitney_u_test}. Given two samples $x = \{x_1, x_2, \dots, x_n\}$ and $y = \{y_1, y_2, \dots, y_m\}$, the Mann-Whitney U-Test Statistic $U$ is defined as:
    \begin{equation}
    U = \sum_{i = 1}^n \sum_{j = 1}^m \text{bool}(x_i < y_j),
    \end{equation}
    where, 
    \begin{align}
    \text{bool}(x_i < y_j) = \begin{cases}1 &\text{if } x_i < y_j,\\0 &\text{otherwise.}\end{cases}
    \label{mask_eqn}
    \end{align}
    The Mann-Whitney U-Test decides whether the distribution of $x$ is the same as that of $y$ based on how much the computed value of $U$ differs from its expected value of $\frac{mn}{2}$ (when values present in $x$ occur almost as frequently as values present in $y$)~\cite{mann_whitney_u_test}. Accordingly, in our case, if a given $x_i$ (MLD of an FP) lies above more than half of the values present in $y = \{y_1, y_2, \dots, y_m\}$, that MLD cannot be regarded as a TP at all (and hence, is an actual FP). Otherwise, it can be regarded as a TP (candidate TP). More formally, an FP can be said to be a Candidate TP when the Mann Whitney U-Test produces a $p$-value greater than 0.05 (for a 95\% CI), i.e., when the null hypothesis is true (the FP lies in the distribution formed by the TPs).
    \item \textbf{Lesion Segmentation}: We use the following set of metrics for performance evaluation: Accuracy (Acc), Prec, Rec, Dice (or F1-Score), IoU, Centerline Dice (clDice)~\cite{cldice}, and Modified Hausdorff Distance (MHD)~\cite{mhd}. clDice is a connectivity-preserving metric that quantifies the extent of spatially contiguous correct pixel predictions by affecting comparisons of segmentation masks with centerline skeletons~\cite{cldice}. MHD involves the computation of the forward ($d(I_{p}, I_{gt})$) and backward ($d(I_{gt}, I_{p})$) mean distances between the two point sets/images $I_{p}$ (prediction) and $I_{gt}$ (ground truth) and returns the largest of the two~\cite{mhd}. It evaluates the correspondence between the complex object boundaries in $I_{p}$ and $I_{gt}$, is robust to small noise perturbations, and at the same time, monotonically increases with the increase in segmentation errors, in contrast to the non-robust and non-monotonic vanilla Hausdorff Distance (HD).
\end{enumerate}

\subsubsection{Implementation Details}

We perform all our experiments using PyTorch~\cite{pytorch} on an NVIDIA A100 GPU with 80GB of internal memory. The specific training settings for the two main stages of ODySSeI are as follows:

\begin{enumerate}
    \item \textbf{Lesion Detection}: We utilize a batch size of 64 with all input images resized to 512 $\times$ 512 and train all our models for 50 epochs while utilizing early stopping to stop training when no improvement is seen on the validation dataset. We select the best performing model on the validation set as measured by a fitness metric ($\mathcal{F} = 0.9 \times \text{mAP@0.50-0.95} + 0.1 \times \text{mAP@0.5}$)~\cite{yolo11} encouraging the best possible precision in localization. In line with the authors of YOLO11~\cite{yolo11}, we utilize a linear learning rate scheduler with a warmup duration of 3 epochs, an initial learning rate of 0.01 for weights and 0.1 for biases (a higher learning rate for biases than weights facilitates faster convergence), and an initial momentum of 0.8. Based on the dataset size used for model training, we use the AdamW optimizer~\cite{adamw} ($\lambda$ = 0.0005, $\beta_1$ = 0.937, and $\beta_1$ = 0.999) for the first 10000 training iterations (facilitates faster initial convergence) and then switch over to the Stochastic Gradient Descent (SGD) with Nesterov Momentum optimizer ($\mu$ = 0.937 and $\lambda$ = 0.0005) for the remaining training iterations (facilitates better fine-tuning and generalizability). An IoU threshold of 0.7 is used for non-max suppression. We use the Ultralytics library~\cite{ultralytics}.

    \item \textbf{Lesion Segmentation}: We utilize a batch size of 16 (maximum possible batch size for segmentation due to GPU constraints) with all input images resized to 256 $\times$ 256 and train all our models for 50 epochs while utilizing early stopping to stop training when no improvement is seen on the validation dataset. We select the best performing model on the validation set as measured by the Dice score. We employ AdamW~\cite{adamw} ($\lambda$ = 0.01, $\beta_1$ = 0.9, and $\beta_1$ = 0.999) as our optimizer and set the initial learning rate to 0.0001. We utilize Cosine Annealing~\cite{cosine_annealing} as the learning rate scheduler.
\end{enumerate}

\section{Author Contributions}

\textbf{AC}: Conceptualization, Methodology, Software, Validation, Formal analysis, Investigation, Data Curation, Visualization, Writing - Original Draft, Writing - Review \& Editing. \textbf{XS}: Conceptualization, Methodology, Writing - Review \& Editing, Supervision. \textbf{TM}: Methodology, Data Curation, Writing - Review \& Editing. \textbf{OS}: Writing - Review \& Editing. \textbf{DA}: Data Curation, Writing - Review \& Editing. \textbf{BDB}: Data Curation, Writing - Review \& Editing. \textbf{SF}: Data Curation, Writing - Review \& Editing. \textbf{OM}: Data Curation, Writing - Review \& Editing. \textbf{EA}: Conceptualization, Writing - Review \& Editing. \textbf{PF}: Conceptualization, Writing - Review \& Editing. \textbf{DT}: Conceptualization, Methodology, Writing - Review \& Editing, Supervision.

\section{Competing Interests}

All authors declare no financial or non-financial competing interests.

\section{Data Availability}

The in-house datasets generated and/or analyzed during the current study are not publicly available due to privacy concerns but are available from the corresponding authors on reasonable request.

\section{Code Availability}

The underlying code for this study is available in GitHub and can be accessed via the following link: \\ \href{https://github.com/LTS4/ODySSeI}{https://github.com/LTS4/ODySSeI}.

\section*{Acknowledgments}
This project is supported by the EPFL Center for Intelligent Systems (CIS) and the EPFL AI Center under the grant ``Next Generation Data Augmentation for Heart Attack Prediction''.
%
%
%

\bibliographystyle{splncs04}
\bibliography{biblio}

@article{
AHA_update,
author = {Seth S. Martin  and Aaron W. Aday  and Norrina B. Allen  and Zaid I. Almarzooq  and Cheryl A.M. Anderson  and Pankaj Arora  and Christy L. Avery  and Carissa M. Baker-Smith  and Nisha Bansal  and Andrea Z. Beaton  and Yvonne Commodore-Mensah  and Maria E. Currie  and Mitchell S.V. Elkind  and Wenjun Fan  and Giuliano Generoso  and Bethany Barone Gibbs  and Debra G. Heard  and Swapnil Hiremath  and Michelle C. Johansen  and Dhruv S. Kazi  and Darae Ko  and Michelle H. Leppert  and Jared W. Magnani  and Erin D. Michos  and Michael E. Mussolino  and Nisha I. Parikh  and Sarah M. Perman  and Mary Rezk-Hanna  and Gregory A. Roth  and Nilay S. Shah  and Mellanie V. Springer  and Marie-Pierre St-Onge  and Evan L. Thacker  and Sarah M. Urbut  and Harriette G.C. Van Spall  and Jenifer H. Voeks  and Seamus P. Whelton  and Nathan D. Wong  and Sally S. Wong  and Kristine Yaffe  and Latha P. Palaniappan  and on behalf of the American Heart Association Council on Epidemiology and Prevention Statistics Committee and Stroke Statistics Committee},
title = {2025 Heart Disease and Stroke Statistics: A Report of US and Global Data From the American Heart Association},
journal = {Circulation},
volume = {151},
number = {8},
pages = {e41-e660},
year = {2025},
doi = {10.1161/CIR.0000000000001303},
URL = {https://www.ahajournals.org/doi/abs/10.1161/CIR.0000000000001303}}

@BOOK{IHD,
  author    = "Institute of Medicine",
  title     = {{Cardiovascular Disability: Updating the Social Security Listings}},
  isbn      = "978-0-309-15698-1",
  year      = 2010,
  publisher = "The National Academies Press",
  address   = "Washington, DC",
  doi       = "10.17226/12940",
  url       = "https://nap.nationalacademies.org/catalog/12940/cardiovascular-disability-updating-the-social-security-listings"
}

@article{CAR_guidelines,
author = {Jacqueline E. Tamis-Holland  and Sripal Bangalore  and Eric R. Bates  and Theresa M. Beckie  and James M. Bischoff  and John A. Bittl  and Mauricio G. Cohen  and J. Michael DiMaio  and Creighton W. Don  and Stephen E. Fremes  and Mario F. Gaudino  and Zachary D. Goldberger  and Michael C. Grant  and Jang B. Jaswal  and Paul A. Kurlansky  and Roxana Mehran  and Thomas S. Metkus  and Lorraine C. Nnacheta  and Sunil V. Rao  and Frank W. Sellke  and Garima Sharma  and Celina M. Yong  and Brittany A. Zwischenberger},
title = {{2021 ACC/AHA/SCAI Guideline for Coronary Artery Revascularization: A Report of the American College of Cardiology/American Heart Association Joint Committee on Clinical Practice Guidelines}},
journal = {Circulation},
volume = {145},
number = {3},
pages = {e18-e114},
year = {2022},
doi = {10.1161/CIR.0000000000001038},
URL = {https://www.ahajournals.org/doi/abs/10.1161/CIR.0000000000001038}}

@article{ICA_variability,
title = {Effect of variability in the interpretation of coronary angiograms on the appropriateness of use of coronary revascularization procedures},
journal = {American Heart Journal},
volume = {139},
number = {1},
pages = {106-113},
year = {2000},
issn = {0002-8703},
author = {Lucian L. Leape and Rolla Edward Park and Thomas M. Bashore and J.Kevin Harrison and Charles J. Davidson and Robert H. Brook},
doi = {https://doi.org/10.1016/S0002-8703(00)90316-8},
url = {https://www.sciencedirect.com/science/article/pii/S0002870300700160}}

@article{QCA,
title = {{Comparison of visual assessment of coronary stenosis with independent quantitative coronary angiography: Findings from the Prospective Multicenter Imaging Study for Evaluation of Chest Pain (PROMISE) trial}},
journal = {American Heart Journal},
volume = {184},
pages = {1-9},
year = {2017},
issn = {0002-8703},
doi = {https://doi.org/10.1016/j.ahj.2016.10.014},
url = {https://www.sciencedirect.com/science/article/pii/S0002870316302368},
author = {Rohan Shah and Eric Yow and William Schuyler Jones and Louis P. Kohl and Andrzej S. Kosinski and Udo Hoffmann and Kerry L. Lee and Christopher B. Fordyce and Daniel B. Mark and Alicia Lowe and Pamela S. Douglas and Manesh R. Patel}
}

@Article{CathAI,
author={Avram, Robert
and Olgin, Jeffrey E.
and Ahmed, Zeeshan
and Verreault-Julien, Louis
and Wan, Alvin
and Barrios, Joshua
and Abreau, Sean
and Wan, Derek
and Gonzalez, Joseph E.
and Tardif, Jean-Claude
and So, Derek Y.
and Soni, Krishan
and Tison, Geoffrey H.},
title={{CathAI: fully automated coronary angiography interpretation and stenosis estimation}},
journal={npj Digital Medicine},
year={2023},
month={Aug},
day={11},
volume={6},
number={1},
pages={142},
issn={2398-6352},
doi={10.1038/s41746-023-00880-1},
url={https://doi.org/10.1038/s41746-023-00880-1}
}

@Article{DeepCoro,
author={Labrecque Langlais, \'Elodie
and Corbin, Denis
and Tastet, Olivier
and Hayek, Ahmad
and Doolub, Gemina
and Mrad, Sebasti\'an
and Tardif, Jean-Claude
and Tanguay, Jean-Fran\ccois
and Marquis-Gravel, Guillaume
and Tison, Geoffrey H.
and Kadoury, Samuel
and Le, William
and Gallo, Richard
and Lesage, Frederic
and Avram, Robert},
title={{Evaluation of stenoses using AI video models applied to coronary angiography}},
journal={npj Digital Medicine},
year={2024},
month={May},
day={23},
volume={7},
number={1},
pages={138},
issn={2398-6352},
doi={10.1038/s41746-024-01134-4},
url={https://doi.org/10.1038/s41746-024-01134-4}
}

@Article{lesion_det2,
AUTHOR = {Wu, Hui and Zhao, Jing and Li, Jiehui and Zeng, Yan and Wu, Weiwei and Zhou, Zhuhuang and Wu, Shuicai and Xu, Liang and Song, Min and Yu, Qibin and Song, Ziwei and Chen, Lin},
TITLE = {One-Stage Detection without Segmentation for Multi-Type Coronary Lesions in Angiography Images Using Deep Learning},
JOURNAL = {Diagnostics},
VOLUME = {13},
YEAR = {2023},
NUMBER = {18},
ARTICLE-NUMBER = {3011},
URL = {https://www.mdpi.com/2075-4418/13/18/3011},
PubMedID = {37761378},
ISSN = {2075-4418},
DOI = {10.3390/diagnostics13183011}
}

@article{lesion_det3,
title = {Training and validation of a deep learning architecture for the automatic analysis of coronary angiography},
author = {Du, Tianming  and Xie, Lihua  and Zhang, Honggang  and Liu, Xuqing  and Wang, Xiaofei  and Chen, Donghao  and Xu, Yang  and Sun, Zhongwei  and Zhou, Wenhui  and Song, Lei  and Guan, Changdong  and Lansky, Alexandra  J.  and Xu, Bo },
year = {2021},
journal = {EuroIntervention},
volume = {17},
number = {1},
pages = {32-40},
doi = {10.4244/EIJ-D-20-00570},
url = {https://doi.org/10.4244/EIJ-D-20-00570},
}

@article{arcade_2,
    title={{StenUNet: Automatic Stenosis Detection from X-ray Coronary Angiography}},
    author={Lin, Hui and Liu, Tom and Katsaggelos, Aggelos and Kline, Adrienne},
    journal={arXiv preprint arXiv:2310.14961},
    year={2023}
}

@article{FAME2_orig,
author = {Bernard De Bruyne  and Nico H.J. Pijls  and Bindu Kalesan  and Emanuele Barbato  and Pim A.L. Tonino  and Zsolt Piroth  and Nikola Jagic  and Sven Möbius-Winkler  and Gilles Rioufol  and Nils Witt  and Petr Kala  and Philip MacCarthy  and Thomas Engström  and Keith G. Oldroyd  and Kreton Mavromatis  and Ganesh Manoharan  and Peter Verlee  and Ole Frobert  and Nick Curzen  and Jane B. Johnson  and Peter Jüni  and William F. Fearon },
title = {{Fractional Flow Reserve–Guided PCI versus Medical Therapy in Stable Coronary Disease}},
journal = {New England Journal of Medicine},
volume = {367},
number = {11},
pages = {991-1001},
year = {2012},
doi = {10.1056/NEJMoa1205361},
URL = {https://www.nejm.org/doi/full/10.1056/NEJMoa1205361}
}

@article{FAME2_new,
author = {Giovanni Ciccarelli  and Emanuele Barbato  and Gabor G. Toth  and Brigitta Gahl  and Panagiotis Xaplanteris  and Stephane Fournier  and Anastasios Milkas  and Jozef Bartunek  and Marc Vanderheyden  and Nico Pijls  and Pim Tonino  and William F. Fearon  and Peter Jüni  and Bernard De Bruyne },
title = {{Angiography Versus Hemodynamics to Predict the Natural History of Coronary Stenoses}},
journal = {Circulation},
volume = {137},
number = {14},
pages = {1475-1485},
year = {2018},
doi = {10.1161/CIRCULATIONAHA.117.028782},
URL = {https://www.ahajournals.org/doi/abs/10.1161/CIRCULATIONAHA.117.028782},
eprint = {https://www.ahajournals.org/doi/pdf/10.1161/CIRCULATIONAHA.117.028782}}

@Article{ARCADE,
author={Popov, Maxim
and Amanturdieva, Akmaral
and Zhaksylyk, Nuren
and Alkanov, Alsabir
and Saniyazbekov, Adilbek
and Aimyshev, Temirgali
and Ismailov, Eldar
and Bulegenov, Ablay
and Kuzhukeyev, Arystan
and Kulanbayeva, Aizhan
and Kalzhanov, Almat
and Temenov, Nurzhan
and Kolesnikov, Alexey
and Sakhov, Orazbek
and Fazli, Siamac},
title={{Dataset for Automatic Region-based Coronary Artery Disease Diagnostics Using X-Ray Angiography Images}},
journal={Scientific Data},
year={2024},
month={Jan},
day={03},
volume={11},
number={1},
pages={20},
issn={2052-4463},
doi={10.1038/s41597-023-02871-z},
url={https://doi.org/10.1038/s41597-023-02871-z}
}

@Article{StenDet,
author={Danilov, Viacheslav V.
and Klyshnikov, Kirill Yu.
and Gerget, Olga M.
and Kutikhin, Anton G.
and Ganyukov, Vladimir I.
and Frangi, Alejandro F.
and Ovcharenko, Evgeny A.},
title={Real-time coronary artery stenosis detection based on modern neural networks},
journal={Scientific Reports},
year={2021},
month={Apr},
day={07},
volume={11},
number={1},
pages={7582},
issn={2045-2322},
doi={10.1038/s41598-021-87174-2},
url={https://doi.org/10.1038/s41598-021-87174-2}
}

@article{Future_Culprit,
author = {Pagnoni, Mattia and Meier, David and Candreva, Alessandro and Maillard, Luc and Adjedj, Julien and Collet, Carlos and Mahendiran, Thabo and Cook, Stephane and Mujcinovic, Alma and Dupré, Marion and Rubimbura, Vladimir and Roguelov, Christan and Eeckhout, Eric and De Bruyne, Bernard and Muller, Olivier and Fournier, Stephane},
title = {{Future culprit detection based on angiography-derived FFR}},
journal = {Catheterization and Cardiovascular Interventions},
volume = {98},
number = {3},
pages = {E388-E394},
doi = {https://doi.org/10.1002/ccd.29736},
url = {https://onlinelibrary.wiley.com/doi/abs/10.1002/ccd.29736},
eprint = {https://onlinelibrary.wiley.com/doi/pdf/10.1002/ccd.29736},
year = {2021}
}

@inproceedings{faster_rcnn,
author = {Ren, Shaoqing and He, Kaiming and Girshick, Ross and Sun, Jian},
title = {{Faster R-CNN: towards real-time object detection with region proposal networks}},
year = {2015},
publisher = {MIT Press},
address = {Cambridge, MA, USA},
booktitle = {Proceedings of the 29th International Conference on Neural Information Processing Systems - Volume 1},
pages = {91–99},
numpages = {9},
location = {Montreal, Canada},
series = {NIPS'15}
}

@INPROCEEDINGS{efficientdet,
  author={Tan, Mingxing and Pang, Ruoming and Le, Quoc V.},
  booktitle={2020 IEEE/CVF Conference on Computer Vision and Pattern Recognition (CVPR)}, 
  title={{EfficientDet: Scalable and Efficient Object Detection}}, 
  year={2020},
  volume={},
  number={},
  pages={10778-10787},
  doi={10.1109/CVPR42600.2020.01079}}

@misc{mosaic,
      title={{YOLOv4: Optimal Speed and Accuracy of Object Detection}}, 
      author={Alexey Bochkovskiy and Chien-Yao Wang and Hong-Yuan Mark Liao},
      year={2020},
      eprint={2004.10934},
      archivePrefix={arXiv},
      primaryClass={cs.CV},
      url={https://arxiv.org/abs/2004.10934}, 
}

@misc{yolo11,
  author = {Glenn Jocher and Jing Qiu},
  title = {{Ultralytics YOLO11}},
  version = {11.0.0},
  year = {2024},
  url = {https://github.com/ultralytics/ultralytics},
  orcid = {0000-0001-5950-6979, 0000-0002-7603-6750, 0000-0003-3783-7069},
  license = {AGPL-3.0}
}

@InProceedings{coco,
author="Lin, Tsung-Yi
and Maire, Michael
and Belongie, Serge
and Hays, James
and Perona, Pietro
and Ramanan, Deva
and Doll{\'a}r, Piotr
and Zitnick, C. Lawrence",
editor="Fleet, David
and Pajdla, Tomas
and Schiele, Bernt
and Tuytelaars, Tinne",
title={{Microsoft COCO: Common Objects in Context}},
booktitle="Computer Vision -- ECCV 2014",
year="2014",
publisher="Springer International Publishing",
address="Cham",
pages="740--755",
}

@article{imagenet,
title = {{ImageNet Large Scale Visual Recognition Challenge}},
author = "Olga Russakovsky and Jia Deng and Hao Su and Jonathan Krause and Sanjeev Satheesh and Sean Ma and Zhiheng Huang and Andrej Karpathy and Aditya Khosla and Michael Bernstein and Berg, {Alexander C.} and Li Fei-Fei",
note = "Publisher Copyright: {\textcopyright} 2015, Springer Science+Business Media New York.",
year = "2015",
month = dec,
day = "1",
doi = "10.1007/s11263-015-0816-y",
language = "English (US)",
volume = "115",
pages = "211--252",
journal = "International Journal of Computer Vision",
issn = "0920-5691",
publisher = "Springer Netherlands",
number = "3",
}

@InProceedings{unet,
author="Ronneberger, Olaf
and Fischer, Philipp
and Brox, Thomas",
editor="Navab, Nassir
and Hornegger, Joachim
and Wells, William M.
and Frangi, Alejandro F.",
title={{U-Net: Convolutional Networks for Biomedical Image Segmentation}},
booktitle="Medical Image Computing and Computer-Assisted Intervention -- MICCAI 2015",
year="2015",
publisher="Springer International Publishing",
address="Cham",
pages="234--241",
isbn="978-3-319-24574-4"
}

@InProceedings{deeplabv3+,
author="Chen, Liang-Chieh
and Zhu, Yukun
and Papandreou, George
and Schroff, Florian
and Adam, Hartwig",
editor="Ferrari, Vittorio
and Hebert, Martial
and Sminchisescu, Cristian
and Weiss, Yair",
title={{Encoder-Decoder with Atrous Separable Convolution for Semantic Image Segmentation}},
booktitle="Computer Vision -- ECCV 2018",
year="2018",
publisher="Springer International Publishing",
address="Cham",
pages="833--851",
isbn="978-3-030-01234-2"
}

@ARTICLE{ciou,
  author={Zheng, Zhaohui and Wang, Ping and Ren, Dongwei and Liu, Wei and Ye, Rongguang and Hu, Qinghua and Zuo, Wangmeng},
  journal={IEEE Transactions on Cybernetics}, 
  title={{Enhancing Geometric Factors in Model Learning and Inference for Object Detection and Instance Segmentation}}, 
  year={2022},
  volume={52},
  number={8},
  pages={8574-8586},
  doi={10.1109/TCYB.2021.3095305}}

@inproceedings{dfl,
author = {Li, Xiang and Wang, Wenhai and Wu, Lijun and Chen, Shuo and Hu, Xiaolin and Li, Jun and Tang, Jinhui and Yang, Jian},
title = {Generalized focal loss: learning qualified and distributed bounding boxes for dense object detection},
year = {2020},
isbn = {9781713829546},
publisher = {Curran Associates Inc.},
address = {Red Hook, NY, USA},
booktitle = {Proceedings of the 34th International Conference on Neural Information Processing Systems},
articleno = {1763},
numpages = {11},
location = {Vancouver, BC, Canada},
series = {NIPS '20}
}

@INPROCEEDINGS{vfl,
author = { Zhang, Haoyang and Wang, Ying and Dayoub, Feras and Sunderhauf, Niko },
booktitle = { 2021 IEEE/CVF Conference on Computer Vision and Pattern Recognition (CVPR) },
title = {{ VarifocalNet: An IoU-aware Dense Object Detector }},
year = {2021},
volume = {},
ISSN = {},
pages = {8510-8519},
doi = {10.1109/CVPR46437.2021.00841},
url = {https://doi.ieeecomputersociety.org/10.1109/CVPR46437.2021.00841},
publisher = {IEEE Computer Society},
address = {Los Alamitos, CA, USA},
month = {Jun}}

@INPROCEEDINGS{soft_dice,
  author={Milletari, Fausto and Navab, Nassir and Ahmadi, Seyed-Ahmad},
  booktitle={2016 Fourth International Conference on 3D Vision (3DV)}, 
  title={{V-Net: Fully Convolutional Neural Networks for Volumetric Medical Image Segmentation}}, 
  year={2016},
  volume={},
  number={},
  pages={565-571},
  doi={10.1109/3DV.2016.79}}

@INPROCEEDINGS{cldice,
  author={Shit, Suprosanna and Paetzold, Johannes C. and Sekuboyina, Anjany and Ezhov, Ivan and Unger, Alexander and Zhylka, Andrey and Pluim, Josien P. W. and Bauer, Ulrich and Menze, Bjoern H.},
  booktitle={2021 IEEE/CVF Conference on Computer Vision and Pattern Recognition (CVPR)}, 
  title={{clDice - a Novel Topology-Preserving Loss Function for Tubular Structure Segmentation}}, 
  year={2021},
  volume={},
  number={},
  pages={16555-16564},
  doi={10.1109/CVPR46437.2021.01629}}

@INPROCEEDINGS{mhd,
  author={Dubuisson, M.-P. and Jain, A.K.},
  booktitle={Proceedings of 12th International Conference on Pattern Recognition}, 
  title={{A modified Hausdorff distance for object matching}}, 
  year={1994},
  volume={1},
  number={},
  pages={566-568 vol.1},
  doi={10.1109/ICPR.1994.576361}}

@article{skeleton,
author = {Zhang, T. Y. and Suen, C. Y.},
title = {A fast parallel algorithm for thinning digital patterns},
year = {1984},
issue_date = {March 1984},
publisher = {Association for Computing Machinery},
address = {New York, NY, USA},
volume = {27},
number = {3},
issn = {0001-0782},
url = {https://doi.org/10.1145/357994.358023},
doi = {10.1145/357994.358023},
journal = {Commun. ACM},
month = mar,
pages = {236–239},
numpages = {4}
}

@article{CLAHE_1,
title = {Adaptive histogram equalization and its variations},
journal = {Computer Vision, Graphics, and Image Processing},
volume = {39},
number = {3},
pages = {355-368},
year = {1987},
issn = {0734-189X},
doi = {https://doi.org/10.1016/S0734-189X(87)80186-X},
url = {https://www.sciencedirect.com/science/article/pii/S0734189X8780186X},
author = {Stephen M. Pizer and E. Philip Amburn and John D. Austin and Robert Cromartie and Ari Geselowitz and Trey Greer and Bart {ter Haar Romeny} and John B. Zimmerman and Karel Zuiderveld}}

@inbook{CLAHE_2,
author = {Zuiderveld, Karel},
title = {Contrast limited adaptive histogram equalization},
year = {1994},
isbn = {0123361559},
publisher = {Academic Press Professional, Inc.},
address = {USA},
booktitle = {Graphics Gems IV},
pages = {474–485},
numpages = {12}
}

@inproceedings{lps,
author = {Zhou, Zongwei and Sodha, Vatsal and Rahman Siddiquee, Md Mahfuzur and Feng, Ruibin and Tajbakhsh, Nima and Gotway, Michael B. and Liang, Jianming},
title = {{Models Genesis: Generic Autodidactic Models for 3D Medical Image Analysis}},
year = {2019},
isbn = {978-3-030-32250-2},
publisher = {Springer-Verlag},
address = {Berlin, Heidelberg},
url = {https://doi.org/10.1007/978-3-030-32251-9_42},
doi = {10.1007/978-3-030-32251-9_42},
booktitle = {Medical Image Computing and Computer Assisted Intervention – MICCAI 2019: 22nd International Conference, Shenzhen, China, October 13–17, 2019, Proceedings, Part IV},
pages = {384–393},
numpages = {10},
location = {Shenzhen, China}
}

@inproceedings{erasing,
title={{Random Erasing Data Augmentation}},
author={Zhong, Zhun and Zheng, Liang and Kang, Guoliang and Li, Shaozi and Yang, Yi},
booktitle={Proceedings of the AAAI Conference on Artificial Intelligence (AAAI)},
year={2020}
}

@Article{albumentations,
    AUTHOR = {Buslaev, Alexander and Iglovikov, Vladimir I. and Khvedchenya, Eugene and Parinov, Alex and Druzhinin, Mikhail and Kalinin, Alexandr A.},
    TITLE = {{Albumentations: Fast and Flexible Image Augmentations}},
    JOURNAL = {Information},
    VOLUME = {11},
    YEAR = {2020},
    NUMBER = {2},
    ARTICLE-NUMBER = {125},
    URL = {https://www.mdpi.com/2078-2489/11/2/125},
    ISSN = {2078-2489},
    DOI = {10.3390/info11020125}
}

@inproceedings{kornia,
  author    = {E. Riba and D. Mishkin and D. Ponsa and E. Rublee and G. Bradski},
  title     = {{Kornia: an Open Source Differentiable Computer Vision Library for PyTorch}},
  booktitle = {Winter Conference on Applications of Computer Vision},
  year      = {2020},
  url       = {https://arxiv.org/pdf/1910.02190.pdf}
}

@inproceedings{pytorch,
 author = {Paszke, Adam and Gross, Sam and Massa, Francisco and Lerer, Adam and Bradbury, James and Chanan, Gregory and Killeen, Trevor and Lin, Zeming and Gimelshein, Natalia and Antiga, Luca and Desmaison, Alban and Kopf, Andreas and Yang, Edward and DeVito, Zachary and Raison, Martin and Tejani, Alykhan and Chilamkurthy, Sasank and Steiner, Benoit and Fang, Lu and Bai, Junjie and Chintala, Soumith},
 booktitle = {Advances in Neural Information Processing Systems},
 editor = {H. Wallach and H. Larochelle and A. Beygelzimer and F. d\textquotesingle Alch\'{e}-Buc and E. Fox and R. Garnett},
 pages = {},
 publisher = {Curran Associates, Inc.},
 title = {PyTorch: An Imperative Style, High-Performance Deep Learning Library},
 url = {https://proceedings.neurips.cc/paper_files/paper/2019/file/bdbca288fee7f92f2bfa9f7012727740-Paper.pdf},
 volume = {32},
 year = {2019}
}

@misc{ultralytics,
author = {Jocher, Glenn and Qiu, Jing and Chaurasia, Ayush},
license = {AGPL-3.0},
month = jan,
title = {{Ultralytics YOLO}},
url = {https://github.com/ultralytics/ultralytics},
version = {8.0.0},
year = {2023}
}

@inproceedings{adamw,
  title={{Decoupled Weight Decay Regularization}},
  author={Ilya Loshchilov and Frank Hutter},
  booktitle={International Conference on Learning Representations},
  year={2017}
}

@inproceedings{
cosine_annealing,
title={{{SGDR}: Stochastic Gradient Descent with Warm Restarts}},
author={Ilya Loshchilov and Frank Hutter},
booktitle={International Conference on Learning Representations},
year={2017},
url={https://openreview.net/forum?id=Skq89Scxx}
}

@article{data_aug_survey,
title = {Data augmentation: A comprehensive survey of modern approaches},
journal = {Array},
volume = {16},
pages = {100258},
year = {2022},
issn = {2590-0056},
doi = {https://doi.org/10.1016/j.array.2022.100258},
url = {https://www.sciencedirect.com/science/article/pii/S2590005622000911},
author = {Alhassan Mumuni and Fuseini Mumuni},
}

@article{angiopy,
	author = {Mahendiran, Thabo and Thanou, Dorina and Senouf, Ortal and Jamaa, Yassine and Fournier, Stephane and De Bruyne, Bernard and Abb{\'e}, Emmanuel and Muller, Olivier and And{\`o}, Edward},
	date = {2025/01/01},
	doi = {10.1016/j.ijcard.2024.132598},
	isbn = {0167-5273},
	journal = {International Journal of Cardiology},
	journal1 = {International Journal of Cardiology},
	publisher = {Elsevier},
	title = {{AngioPy Segmentation: An open-source, user-guided deep learning tool for coronary artery segmentation}},
	type = {doi: 10.1016/j.ijcard.2024.132598},
	url = {https://doi.org/10.1016/j.ijcard.2024.132598},
	volume = {418},
	year = {2025},
	year1 = {2025}}

@article{mann_whitney_u_test,
author = {H. B. Mann and D. R. Whitney},
title = {{On a Test of Whether one of Two Random Variables is Stochastically Larger than the Other}},
volume = {18},
journal = {The Annals of Mathematical Statistics},
number = {1},
publisher = {Institute of Mathematical Statistics},
pages = {50 -- 60},
year = {1947},
doi = {10.1214/aoms/1177730491},
URL = {https://doi.org/10.1214/aoms/1177730491}
}

\end{document}